\documentclass[final]{cvpr}

\usepackage{times}
\usepackage{epsfig}
\usepackage{graphicx}
\usepackage{amsmath}
\usepackage{amssymb}

\usepackage{multirow}
\usepackage[noend]{algpseudocode}
\usepackage{enumitem}
\usepackage{caption} 
\usepackage{graphicx}
\usepackage{caption}
\usepackage{subcaption}

\usepackage[ruled,vlined]{algorithm2e}
\usepackage{booktabs}
\usepackage{siunitx}

\usepackage[pagebackref=true,breaklinks=true,colorlinks,bookmarks=false]{hyperref}

\def\cvprPaperID{7719} 
\def\confYear{CVPR 2021}

\begin{document}

\title{Joint Negative and Positive Learning for Noisy Labels}

\author{Youngdong Kim ~~~~~~~~~~~~~ Juseung Yun ~~~~~~~~~~~~~ Hyounguk Shon ~~~~~~~~~~~~~ Junmo Kim\\
School of Electrical Engineering, KAIST, South Korea \\
{\tt\small \{ydkim1293, juseung\textunderscore yun, hyounguk.shon, junmo.kim\}@kaist.ac.kr}
}


\maketitle
\thispagestyle{empty}

\begin{abstract}
Training of Convolutional Neural Networks (CNNs) with data with noisy labels is known to be a challenge. Based on the fact that directly providing the label to the data (Positive Learning; PL) has a risk of allowing CNNs to memorize the contaminated labels for the case of noisy data, the indirect learning approach that uses complementary labels (Negative Learning for Noisy Labels; NLNL) has proven to be highly effective in preventing overfitting to noisy data as it reduces the risk of providing faulty target. NLNL further employs a three-stage pipeline to improve convergence. As a result, filtering noisy data through the NLNL pipeline is cumbersome, increasing the training cost. In this study, we propose a novel improvement of NLNL, named Joint Negative and Positive Learning (JNPL), that unifies the filtering pipeline into a single stage. JNPL trains CNN via two losses, NL+ and PL+, which are improved upon NL and PL loss functions, respectively. We analyze the fundamental issue of NL loss function and develop new NL+ loss function producing gradient that enhances the convergence of noisy data. Furthermore, PL+ loss function is designed to enable faster convergence to expected-to-be-clean data. We show that the NL+ and PL+ train CNN simultaneously, significantly simplifying the pipeline, allowing greater ease of practical use compared to NLNL. With a simple semi-supervised training technique, our method achieves state-of-the-art accuracy for noisy data classification based on the superior filtering ability.

\end{abstract}

\begin{figure*}[ht]

\begin{center}
\begin{tabular}{cc}
\hspace*{-0.4cm}
\includegraphics[height=2.85cm]{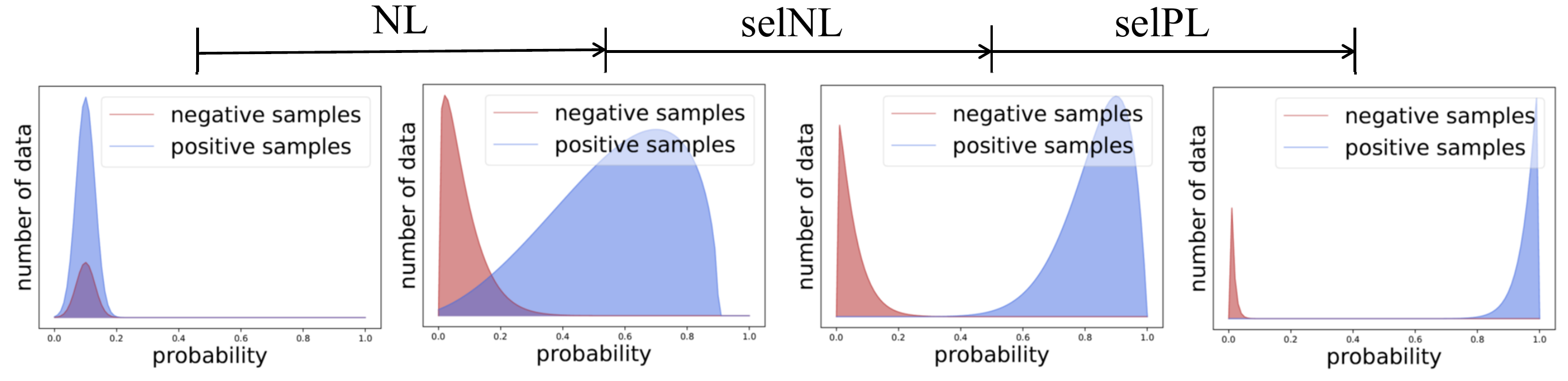} & \includegraphics[height=2.85cm]{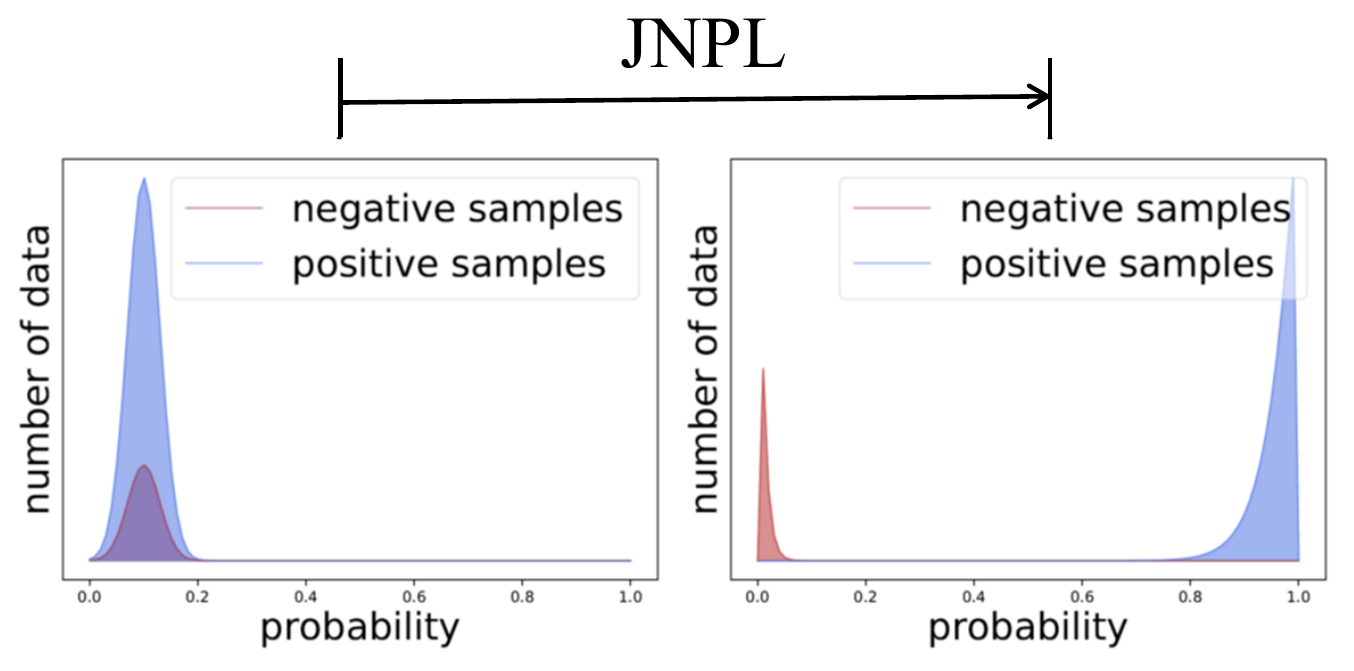} \\ 
(a) NLNL & (b) JNPL \\
\end{tabular}
\end{center}
\caption{Comparison between Negative Learning for Noisy Labels (NLNL) and Joint Negative and Positive Learning (JNPL) for filtering noisy data from training data, demonstrated with histograms showing the distribution of noisy training data . (a): NLNL is a 3-stage pipeline (NL$\rightarrow$SelNL$\rightarrow$SelPL). (b): JNPL is a single-stage pipeline, in which two loss functions (NL+ and PL+) train CNN simultaneously.}
\label{fig:conceptual}
\vspace{-5mm}
\end{figure*}

\section{Introduction}
\label{sec:Introduction}

Convolutional Neural Networks (CNNs) have led to great improvements in many supervised tasks. However, CNNs' performance relies heavily on the quality of labels, and accurately labeling a huge amount of data is expensive and time-consuming. Furthermore, accurate labeling is done by hand, which can eventually lead to mismatched labeling. Therefore, the robust training of CNNs with noisy data is of great practical importance. 
There are many approaches regarding this issue. 
For example, there are methods that design noise-robust loss~\cite{ghosh2015making, ghosh2017robust,wang2019symmetric,ma2020normalized}, use two neural networks to select clean labels~\cite{han2018co,yu2019does,wei2020combating}, and utilize label correction~\cite{patrini2017making,xiao2015learning}.
These existing approaches commonly use the given labels in a direct manner, i.e., ``input image belongs to this label'' (\textit{Positive Learning; PL}). This behavior carries the risk of providing faulty information to the CNNs when noisy labels are involved. 

Motivated by this reason, \textit{Negative Learning for Noisy Labels; NLNL}~\cite{kim2019nlnl}, which is an indirect learning method for training CNNs, has been proposed recently. \textit{Negative Learning} (NL) uses randomly chosen complementary labels and trains the CNN that ``input image does not belong to this complementary label,'' reducing the risk of providing the wrong information because of the high chance of not selecting a true label as a complementary label. Additionally, NLNL proposed three-stage pipeline for filtering noisy data from training data (Figure~\ref{fig:conceptual} (a)). Each stage is composed of NL $\rightarrow$ NL while discarding data of low confidence (\textit{Selective NL; SelNL}) $\rightarrow$ PL while only retaining data of high confidence (\textit{Selective PL; SelPL}), enabling more convergence after NL. However, the fundamental problem that NL loss function causes underfitting to the overall training data still remains. This is the reason that NL requires an additional sequential step, SelNL. Furthermore, the three-stage pipeline for filtering noisy data is quite inefficient, extending the time for training CNNs.

In this study, we propose a novel version of NLNL: \textit{Joint Negative Learning and Positive Learning; JNPL} which has a unified single-stage pipeline for filtering noisy data (Figure~\ref{fig:conceptual} (b)). JNPL is composed of two losses to train CNN, NL+ and PL+ losses, dedicated to filtering noisy data from training data. Each is developed from NL and PL loss functions, respectively. Firstly, our paper focuses on analyzing the NL loss function to understand the cause for underfitting. Then we develop a new loss function NL+ that resolves the issue, which produces a gradient appropriate for convergence on a noisy training dataset. Our study demonstrates the effectiveness of NL+, showing improved convergence across various label noise types and noise rates. Secondly, while we utilize PL to aid in training with noisy data, PL+ loss function is also newly designed to enable faster training with expected-to-be-clean data. Our paper shows the effectiveness of the PL+ loss function compared to the previous PL loss function. Finally, as both loss functions of our method (NL+ and PL+) jointly train the model through a \textit{single stage}, it is simple and easier to use than NLNL. Our experiments show that JNPL successfully filters noisy data in a single stage, thereby providing significantly faster training of CNN as well as better filtering compared to NLNL.

After filtering noisy data from the training data we perform pseudo-labeling for noisy data classification. We achieve state-of-the-art accuracy across various settings in CIFAR10, CIFAR100~\cite{cifar-10}, and Clothing1M~\cite{xiao2015learning} datasets, proving the superior filtering ability of JNPL.

The main contributions of this paper are as follows:

\begin{description}[font=$\bullet$\scshape\bfseries, leftmargin=4mm, topsep=1mm, noitemsep]

  \item We propose an improved version of NLNL, named \textit{``Joint Negative and Positive Learning (JNPL),''} featuring a single-stage pipeline for filtering noisy data, therefore enabling easier usage compared to NLNL.
  \item Two novel loss functions are newly designed, each named NL+ loss and PL+ loss. NL+ solves the underfitting problem of the NL loss, and provides better convergence on various types and ratios of label noises in the training data. Moreover, PL+ enables faster training compared to the previous PL loss function.
  \item Our method filters noisy data, more robust across different types and ratios of noise than NLNL. Our method also achieves state-of-the-art noisy data classification results when used along with pseudo-labeling.
  \item Prior knowledge of the type or number of noisy data is not required for our method. It does not require any hyper-parameter tuning that depend on prior knowledge, allowing our method to be applicable in practice.
\end{description}

The remainder of this paper is organized as follows. Section~\ref{sec:NLNL} describes NLNL method in depth, which is targeted throughout the whole paper, and discusses the cause of the underfitting problem of the method. Section~\ref{sec:JNPL} describes our proposed method, JNPL, and explains in detail on NL+ loss and PL+ loss terms. Section~\ref{sec:Analysis} demonstrates the overall comparison between JNPL and NLNL, showing the distinct advantages of JNPL over NLNL. Section~\ref{sec:Experiments} discusses the evaluations of our method in comparison to baseline methods. Finally, we summarize and conclude in Section~\ref{sec:Conclusion}.

\section{Related works}
\label{sec:Related works}
Several methods that aim to mitigate label noise have been proposed. Here, we summarize some of the recent approaches to noise-robust learning.

\noindent\textbf{Designing noise-robust loss} The commonly used cross-entropy (CE) loss is known to be prone to overfitting when there is noise in the labels. Therefore, a family of studies aims to design novel loss functions that are tolerant of label noise. Ghosh \etal~\cite{ghosh2015making, ghosh2017robust} showed that the mean absolute error (MAE) loss is theoretically robust against label noise. Zhang \etal~\cite{zhang2018generalized} proposed Generalized Cross Entropy loss, which is a generalized function that can interpolate between the forms of CE and MAE, which enables it to adjust trade-offs between robust loss and non-robust loss.

However, in many cases, such noise-robust losses carry the problem of underfitting, which motivates the combination of a robust loss with a non-robust loss to improve convergence. 
Wang \etal~\cite{wang2019symmetric} proposed Symmetric Cross Entropy loss, which combines CE loss with Reverse Cross Entropy loss.
Recently, Ma \etal~\cite{ma2020normalized} proposed a loss normalization technique that transforms a non-robust loss function into a robust loss function. They also showed that such normalized loss used in combination with another robust loss function improves convergence and coined the term Active Passive Loss (APL). 

\noindent\textbf{Weighting samples} In some researches, each sample in the training set is weighted by the reliability of the label~\cite{jiang2017mentornet,ren2018learning,lee2017cleannet}. Moreover, other methods proposed meta-learning algorithms that predicts the weights for each sample~\cite{jiang2017mentornet,ren2018learning}. However, these methods require a clean validation set, which is often difficult to guarantee in practice. 

\noindent\textbf{Correction methods} 
Some other researches used correction methods~\cite{patrini2016making,vahdat2017toward,hendrycks2018using,xiao2015learning,veit2017learning, li2017learning}. They assume that prior knowledge like noise rate or noisy transition matrix is known or that some clean data is accessible. However, in a practical case, prior knowledge and clean data is usually hard to obtain. Some other works used CNN with additional layer~\cite{sukhbaatar2014training, jindal2016learning, goldberger2016training}, and noise transition matrix is approximated to correct loss.
Many efforts gradually change the data label to the prediction value of the network~\cite{reed2014training,tanaka2018joint,ma2018dimensionality,yi2019probabilistic}.
Arazo \etal \cite{arazo2019unsupervised}, fits a mixture of beta distributions that models the loss of clean and noisy samples during training.

\noindent\textbf{Selecting clean labels} 
Some attempted to identify clean labels from a noisy dataset~\cite{han2018co,ding2018semi,northcutt2017learning}. Ding \etal~\cite{ding2018semi} proposed a selection of clean examples based on predicted likelihoods. The labels of the remaining samples are discarded, and the network is trained by semi-supervision. Some of the successful approaches train two deep neural networks simultaneously and let them teach each other \cite{han2018co,yu2019does,wei2020combating}. Each network selects possibly clean data and trains the other network with this data.

\noindent\textbf{Use of complementary labels} Kim \etal~\cite{kim2019nlnl} proposed a noise-robust learning method where instead of maximizing the log-likelihood on the target position, it minimizes the log-likelihood on the complementary positions, termed Negative Learning (NL). They employ a three-stage pipeline based on NL that separates the clean data from the noisy data. Finally, the network is trained using standard CE loss with semi-supervision by treating the noisy set as unlabeled.

\noindent\textbf{Other approaches} Li \etal \cite{li2019learning} uses meta-learning to obtain weights that can be easily fine-tuned to a given noisy dataset. Zhang \etal \cite{zhang2019metacleaner} proposed to learn confidence scores of each samples from the relationship between noisy samples in the feature space, then use the confidence scores to generate cleaner representations. Harutyunyan \etal \cite{harutyunyan2020improving} proposed training algorithm based on mutual information between weights and labels to regularize the memorization of labels.
\section{Negative Learning for Noisy Labels (NLNL)}
\label{sec:NLNL}
Throughout this paper, we consider the problem of c-class classification. Let \( \boldsymbol{x} \in \mathcal{X} \) be an input, \( y, \overline{y} \in \mathcal{Y} = \{1,...,c\}\) be its label and complementary label, respectively, and \( \boldsymbol{y}, \boldsymbol{\overline{y}} \in \{0,1\}^{c}\) be their one-hot vector. Suppose the CNN $f(x;\theta)$ maps the input space to the c-dimensional score space \( f : \mathcal{X} \rightarrow \mathbb{R}^c \), where $\theta$ is the set of network parameters. 
If $f$ passes through the softmax function, the output can be interpreted as a probability vector \(\boldsymbol{p} \in \Delta^{c-1} \), where \(\Delta^{c-1}\) denotes the c-dimensional simplex.

NL~\cite{kim2019nlnl} is an indirect learning method for training CNNs with noisy data. Instead of using given labels, it chooses random complementary label $\overline{y}$ and train CNNs as in ``input image does not belong to this complementary label.'' The loss function following this definition is as below, along with the classic PL loss function for comparison:

\begin{equation}
\label{eq:PL}
\mathcal{L}_{PL}(f,y) = -\sum_{k=1}^c \boldsymbol{y}_k \log \boldsymbol{p}_k
\end{equation}

\begin{equation}
\label{eq:NL}
\mathcal{L}_{NL}(f,\overline{y}) = -\sum_{k=1}^c \boldsymbol{\overline{y}}_k \log(1-\boldsymbol{p}_k).
\end{equation}

To improve convergence after NL, SelNL is performed as a subsequent step. SelNL trains the CNNs only with the data having confidence over $\frac{1}{c}$ ($\boldsymbol{p}_{y}>\frac{1}{c}$). Since data involved in training tend to be less noisy than before, CNNs converge better after SelNL. Furthermore, PL is considered a faster and more accurate method than NL, only if training data is assumed to be clean. After training with NL and SelNL, SelPL train CNNs only with data that has confidence above $\gamma$ $(=0.5)$, assuming that such data are clean. After filtering noisy data with these three steps (NL$\rightarrow$SelNL$\rightarrow$SelPL), semi-supervised learning (pseudo-labeling~\cite{lee2013pseudo}) is performed utilizing labeled expected-to-be-clean data and unlabeled noisy data.

\begin{figure*}[t]
\begin{tabular}{cccc}
\includegraphics[width=4cm]{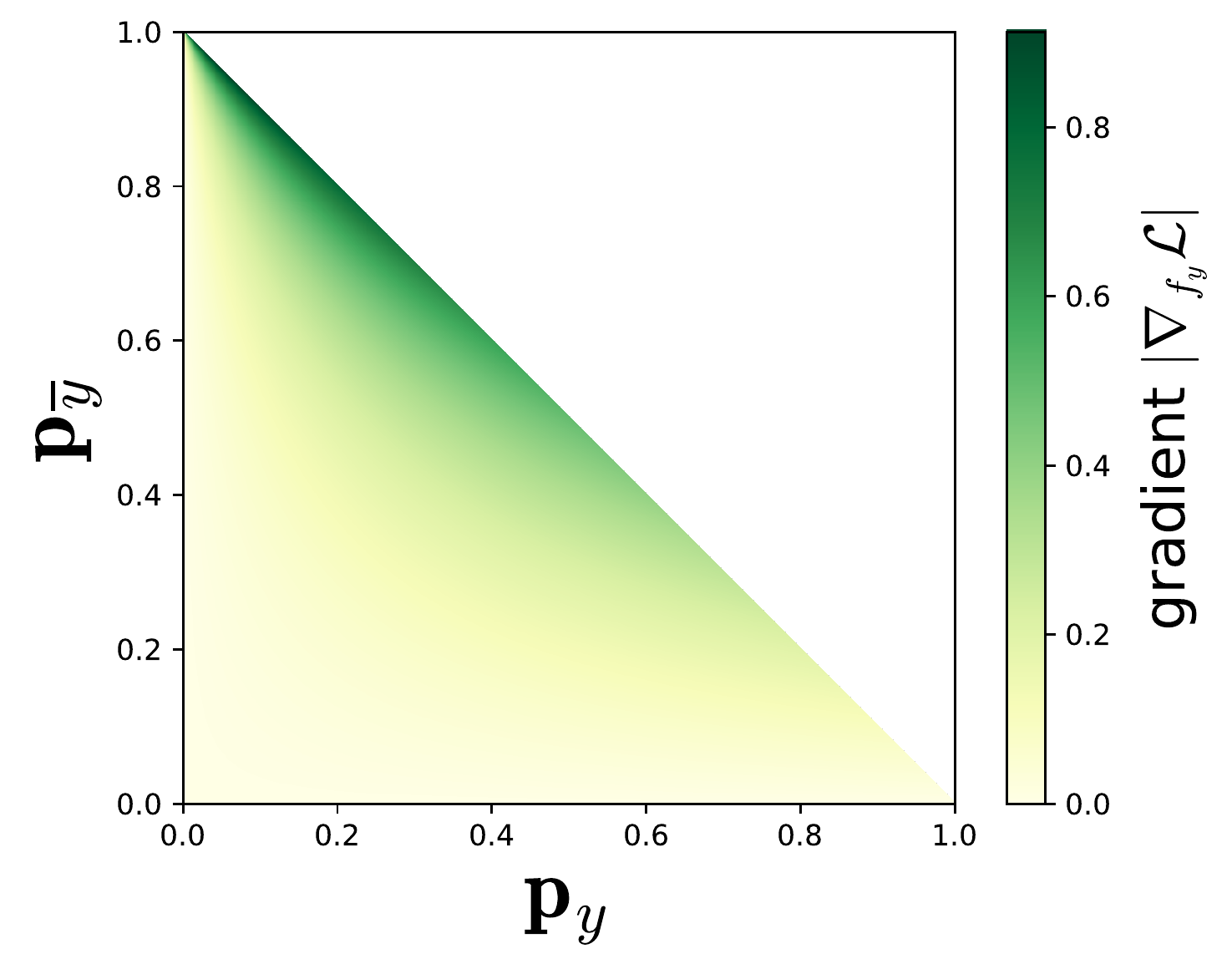}&\includegraphics[width=4cm]{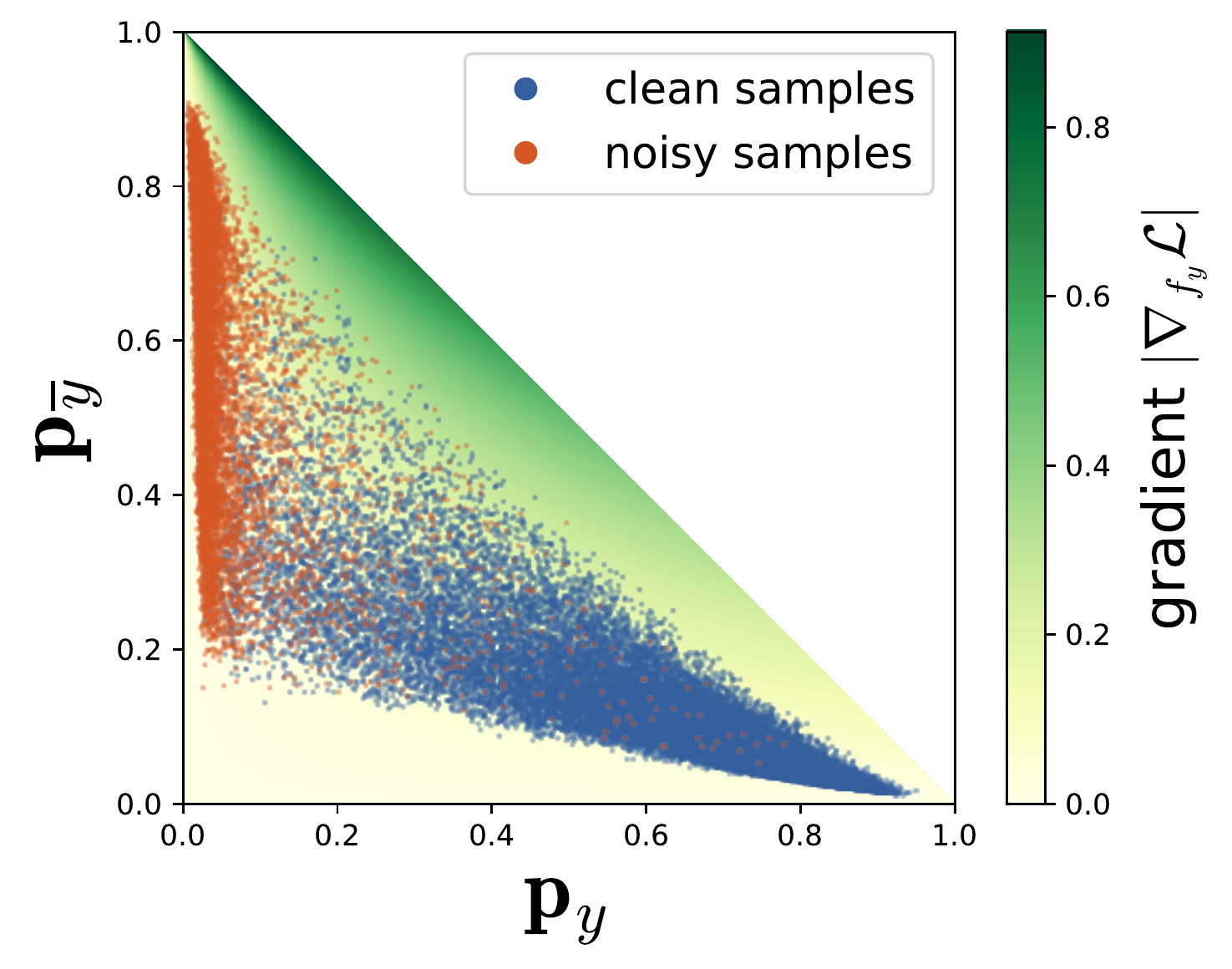}&\includegraphics[width=4cm]{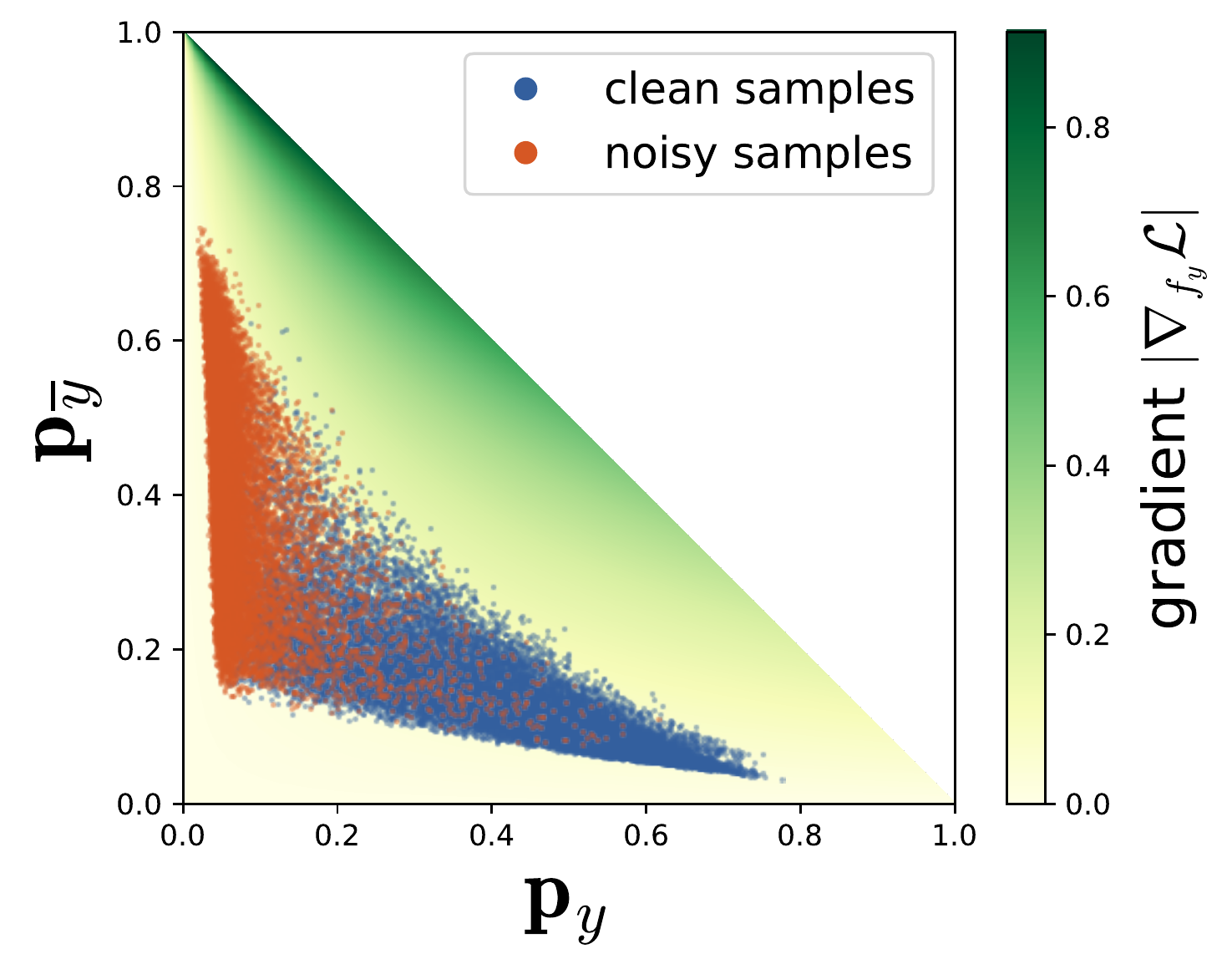}&\includegraphics[width=4cm]{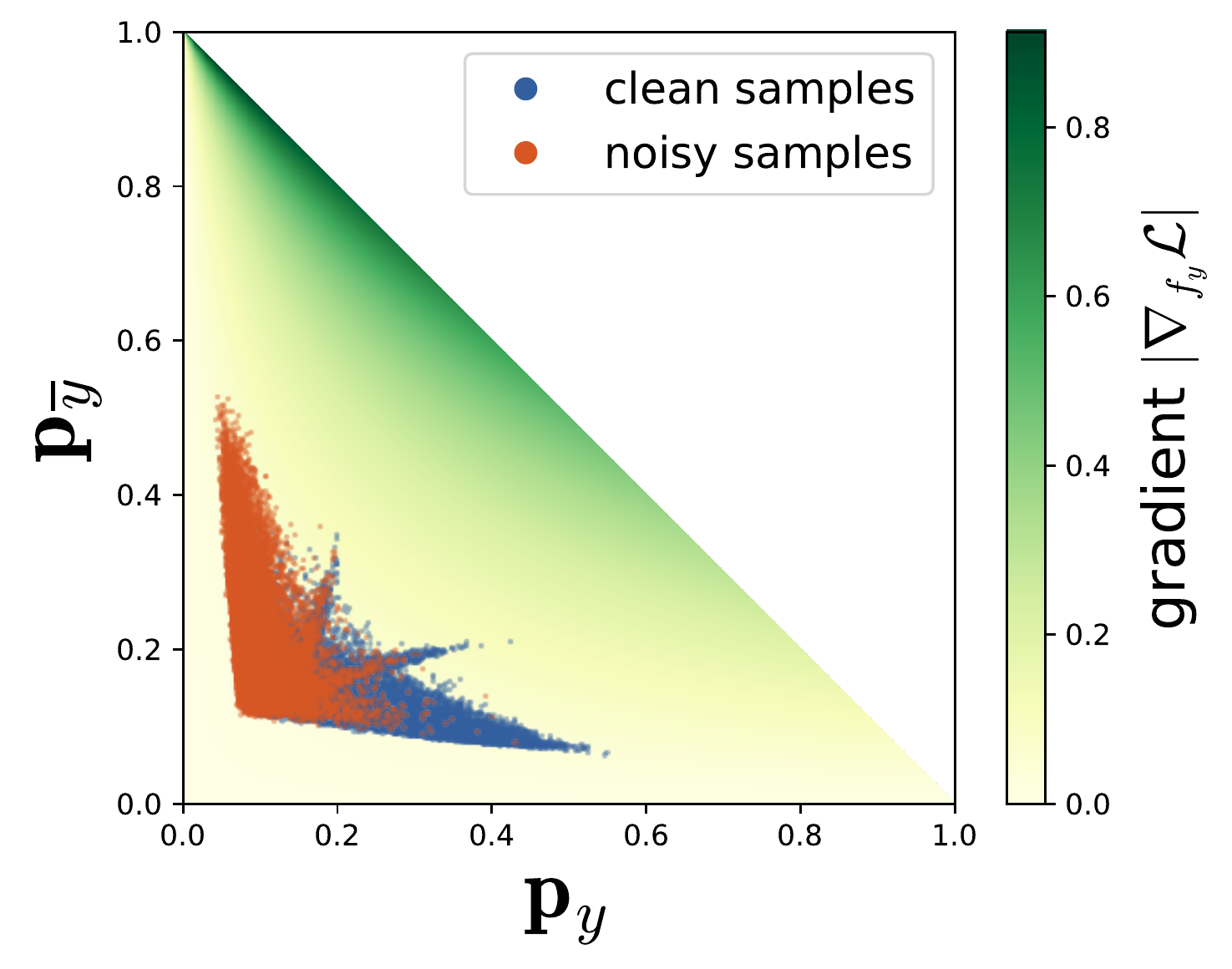}\\
(a) Plot of $\nabla\mathcal{L}_{NL}$ & (b) $\eta=0.2$ & (c) $\eta=0.4$ & (d) $\eta=0.6$\\
\\
\includegraphics[width=4cm]{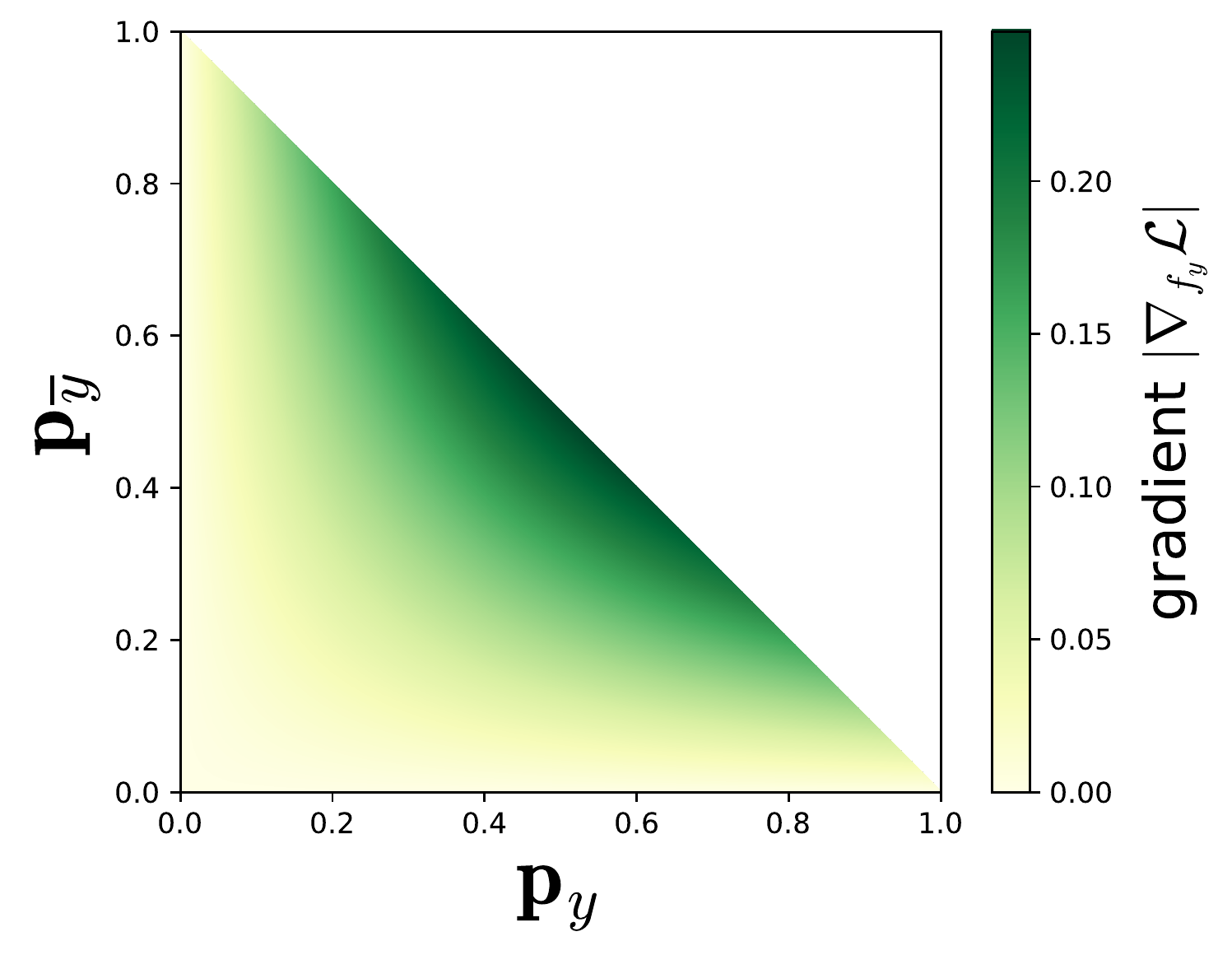}&\includegraphics[width=4cm]{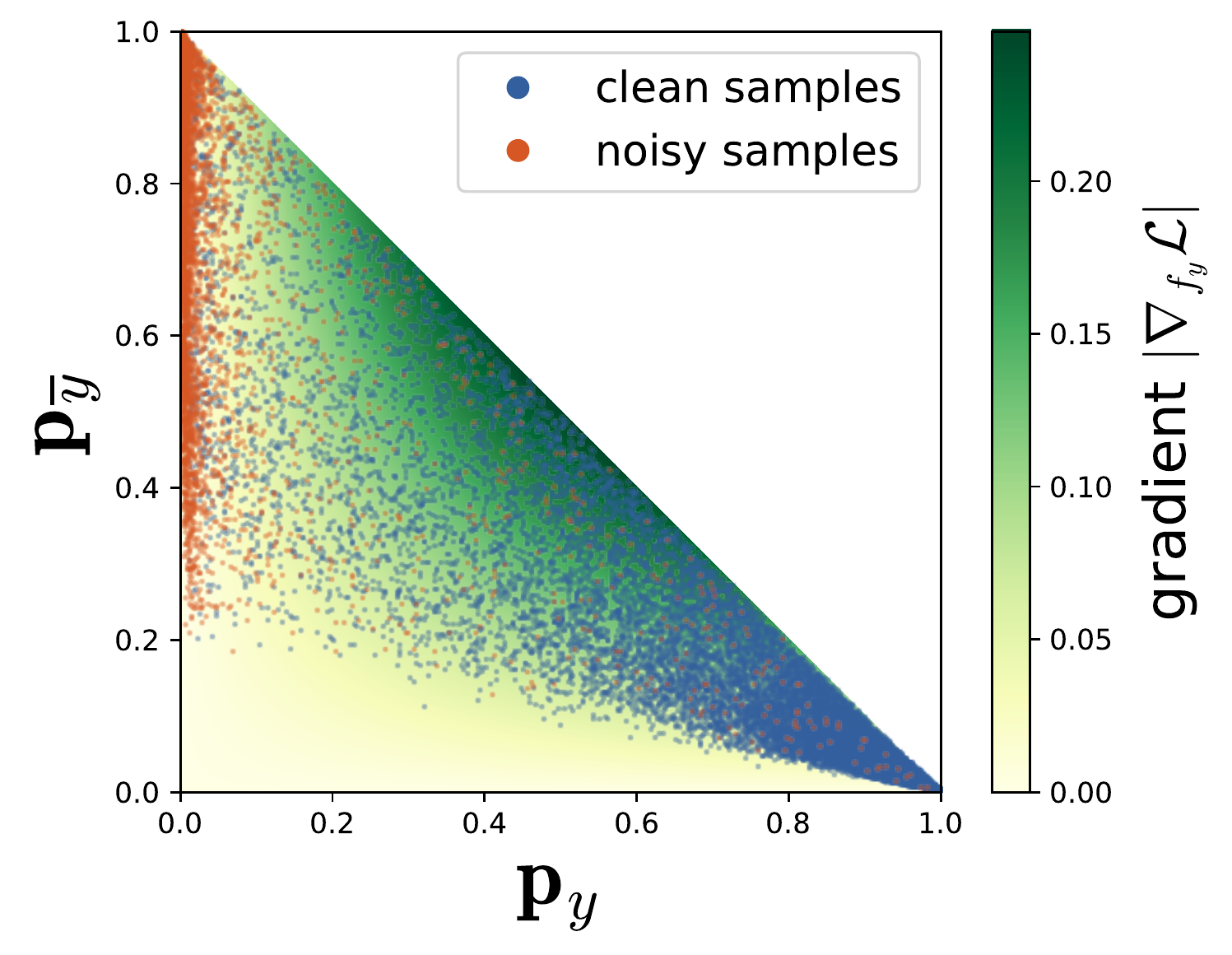}&\includegraphics[width=4cm]{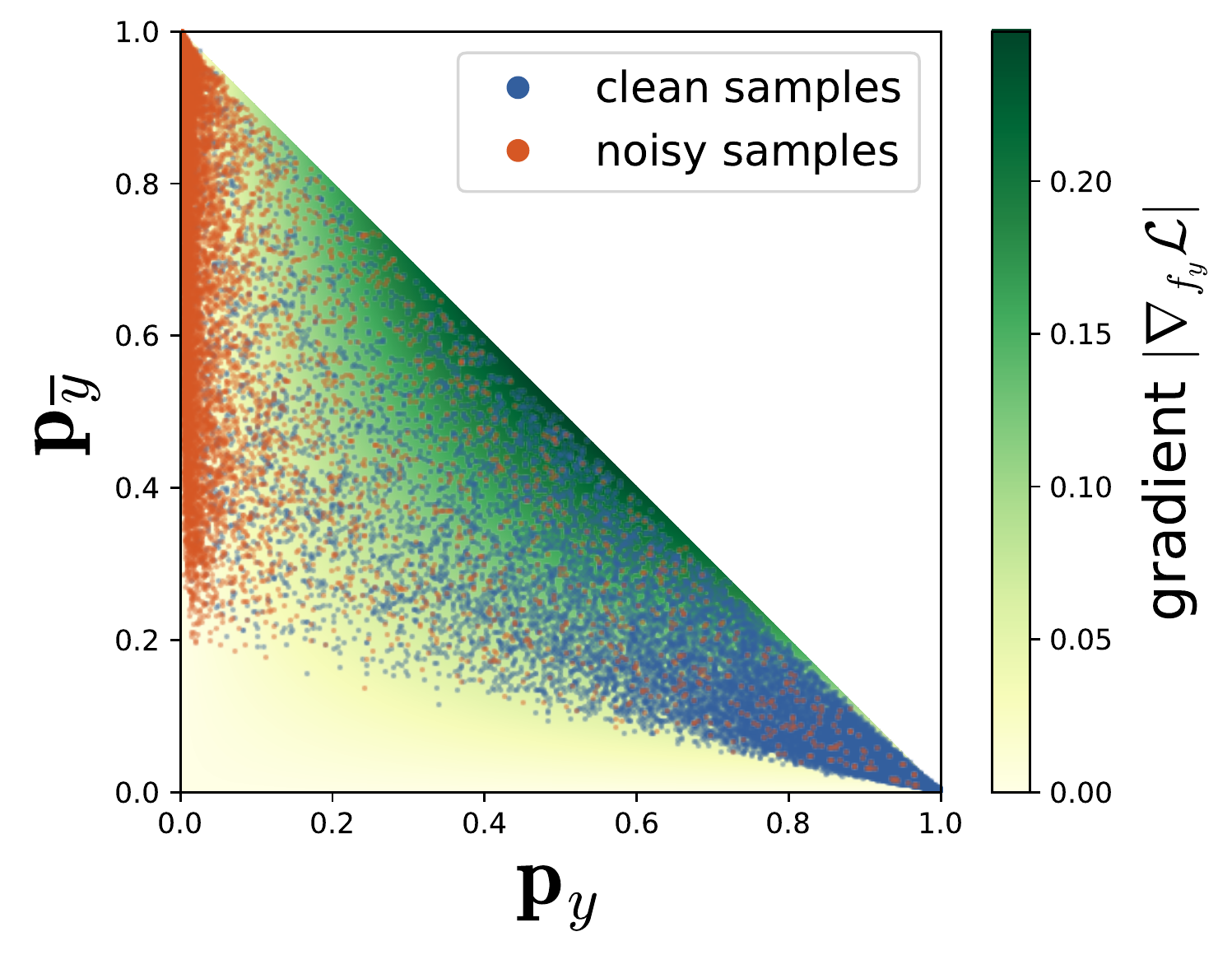}&\includegraphics[width=4cm]{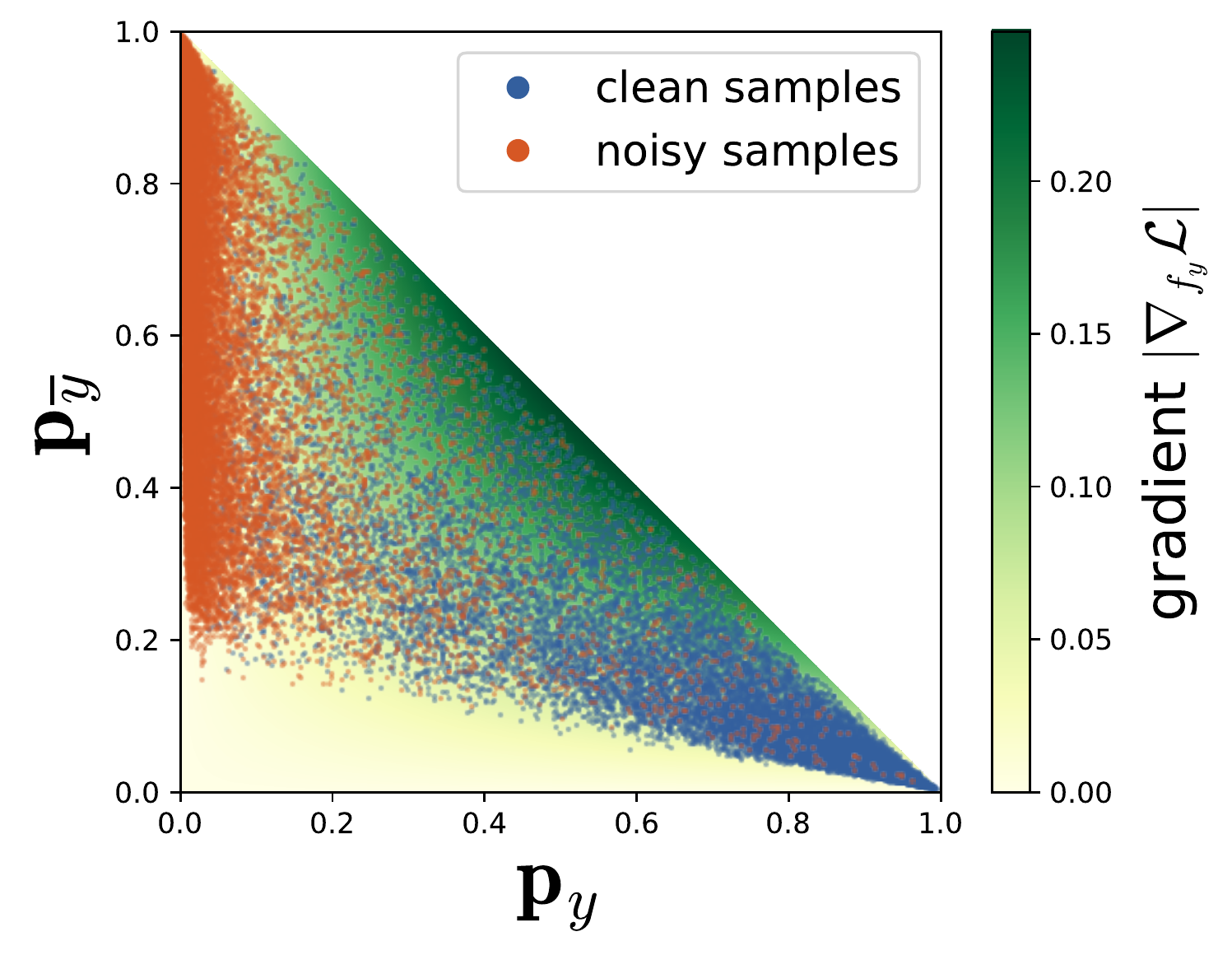}\\
(e) Plot of $\nabla\mathcal{L}_{NL+}$ & (f) $\eta=0.2$ & (g) $\eta=0.4$ & (h) $\eta=0.6$
\end{tabular}
\caption{Comparison between NL and NL+ with CIFAR10 with \textit{symm} noise. (a), (e): Gradient map of NL and NL+, respectively. (b)-(d): Training data distribution with 20\%, 40\%, 60\% noise after training with NL. (f)-(h): Training data distribution with 20\%, 40\%, 60\% noise after training with NL+}
\label{fig:NL_NL+_gradient}
\vspace{-4mm}
\end{figure*}

As mentioned in Section~\ref{sec:Introduction}, the fundamental problem of underfitting of NL still remains. To analyze the root of this phenomenon, we observe the gradient resulting from the NL loss function (Eq~\ref{eq:NL}) as follows:

\begin{equation}
\label{eq:NL gradient}
\begin{split}
\nabla{\mathcal{L}_{NL}}=\frac{\partial \mathcal{L}_{NL}(f,\overline{y})}{\partial f_i} =
\begin{cases}
    \boldsymbol{p}_i & \text{if $i=\overline{y}$} \\
    -\frac{\boldsymbol{p}_{\overline{y}}}{1-\boldsymbol{p}_{\overline{y}}}\boldsymbol{p}_i & \text{if $i \neq \overline{y}.$}
\end{cases}
\end{split}
\end{equation}
Eq~\ref{eq:NL gradient} states that at classes except for $\overline{y}$ receives gradient of $-\frac{\boldsymbol{p}_{\overline{y}}}{1-\boldsymbol{p}_{\overline{y}}}\boldsymbol{p}_i$ ($\nabla{\mathcal{L}_{NL(i\neq{\overline{y}})}}$). Figure~\ref{fig:NL_NL+_gradient} (a) shows 2D gradient map of $\nabla{\mathcal{L}_{NL(i\neq{\overline{y}})}}$, and Figure~\ref{fig:NL_NL+_gradient} (b)-(d) shows the distribution of training data after NL in diverse noise ratio. Each training data is distributed in gradient map with respect to its $\boldsymbol{p}_y$ (when $i=y$) and $\boldsymbol{p}_{\overline{y}_{max}}$. As the training with NL progresses, clean data tend to have high $\boldsymbol{p}_{y}$ and low $\boldsymbol{p}_{\overline{y}}$ (lower-right region in Figure~\ref{fig:NL_NL+_gradient} (a)), while noisy data tend to have low $\boldsymbol{p}_{y}$ and high $\boldsymbol{p}_{\overline{y}}$ (upper-left region in Figure~\ref{fig:NL_NL+_gradient} (a)). However, considering noisy data, ground-truth labels may be chosen as $\overline{y}$. In this case, all classes, except for ground truth label, receive high $\nabla{\mathcal{L}_{NL(i\neq{\overline{y}})}}$ because of high $\boldsymbol{p}_{\overline{y}}$, resulting in underfitting of that data as the confidence of classes other than the ground-truth label increases. In Section~\ref{sec:NL+}, we describe the developed loss function of NL (NL+) that resolves this underfitting issue.

\section{Joint Negative and Positive Learning (JNPL)}
\label{sec:JNPL}
The loss function of the proposed method, JNPL, is composed of two loss functions:

\begin{equation}
\label{eq:JNPL}
\mathcal{L}_{JNPL}=\mathcal{L}_{NL+}+\lambda\mathcal{L}_{PL+}.
\end{equation}
Each of which is dedicated to filtering noisy data from training data. $\mathcal{L}_{NL+}$ is the advanced version of NL, which resolves underfitting issue. $\mathcal{L}_{PL+}$ is other newly designed loss for PL that trains on expected-to-be-clean data, empowering training on data of higher confidence. $\lambda$ is added to scale the overall magnitude of PL+ so that it does not overwhelm the magnitude of NL+. We set $\lambda=0.01$ throughout the whole paper. These two losses enable successful filtering of noisy data. Finally, noisy data classification is done in semi-supervised manner, utilizing these filtered noisy data confidence as pseudo-label. In the following sections, we further introduce each of the loss functions and describe the concept and implementation respectively.

\subsection{NL+}
\label{sec:NL+}

As discussed in Section~\ref{sec:NLNL}, we argue that the cause of the underfitting problem with NL is due to the nature of its gradient $\nabla{\mathcal{L}_{NL(i\neq{\overline{y}})}}$ (Figure~\ref{fig:NL_NL+_gradient} (a)). This is more pronounced as the noise rate increases, as shown in Figure~\ref{fig:NL_NL+_gradient} (b)-(d). This problem occurs when noisy data receives high gradient to classes except for $\overline{y}$ when the confidence of $\overline{y}$ is high, $\overline{y}$ being most likely to be ground truth label. To solve this issue, we propose a modification to the NL loss function, named NL+ loss, as follows:

\begin{equation}
\label{eq:NL+}
\mathcal{L}_{NL+}(f,\overline{y}) = -(1-\boldsymbol{p}_{\overline{y}})\sum_{k=1}^c \boldsymbol{\overline{y}}_k \log(1-\boldsymbol{p}_k).
\end{equation}
It should be noted that $\left(1-\boldsymbol{p}_{\overline{y}}\right)$ acts as a constant weighting factor. Intuitively, this factor has the effect of decreasing the loss for noisy data when corresponding $\boldsymbol{p}_{\overline{y}}$ is high, $\overline{y}$ being most likely to be ground truth label. That way, it reduces the risk of pressing down on the confidence of ground truth label for noisy data, reducing the risk of underfitting. This is further analyzed by observing the gradient of NL+ ($\nabla{\mathcal{L}_{NL+(i\neq{\overline{y}})}}$), given by Eq~\ref{eq:NL+}:

\begin{equation}
\label{eq:NL+ gradient}
\nabla{\mathcal{L}_{NL+(i\neq{\overline{y}})}}=(1-\boldsymbol{p}_{\overline{y}})\nabla{\mathcal{L}_{NL(i\neq{\overline{y}})}}=-\boldsymbol{p}_{\overline{y}}\boldsymbol{p}_i.
\end{equation}
The gradient map of $\nabla{\mathcal{L}_{NL+(i\neq{\overline{y}})}}$ is shown in Figure~\ref{fig:NL_NL+_gradient} (e). Compared to Figure~\ref{fig:NL_NL+_gradient} (a), it shows gradient at upper-left region is reduced. This implies that as the training progresses with NL+, noisy data is gathered at the upper-left region. With NL+, gradient received for noisy data of high $\boldsymbol{p}_{\overline{y}}$ is reduced, allowing noisy data to maintain high $\boldsymbol{p}_{\overline{y}}$ value, where $\overline{y}$ is most likely to be ground truth label. Figure~\ref{fig:NL_NL+_gradient} (f)-(h) shows the distribution of training data mixed with diverse ratio of noise. It shows that compared to Figure~\ref{fig:NL_NL+_gradient} (b)-(d), NL+ results in more convergence. Especially in noise of high ratio (Figure~\ref{fig:NL_NL+_gradient} (d), (h)), NL+ successfully divides noisy data from training data, sending noisy data to upper-left region. 

\begin{figure}[t]
\begin{tabular}{cc}
\includegraphics[width=3cm]{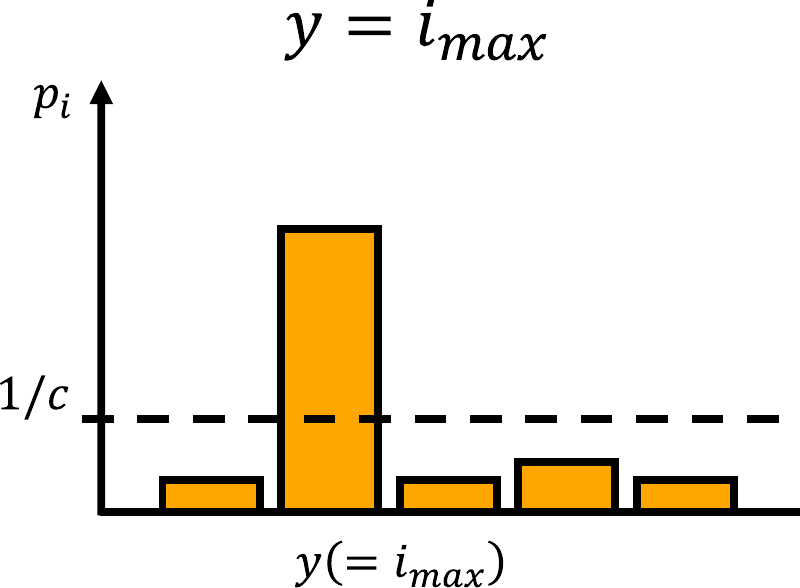} & \includegraphics[width=3cm]{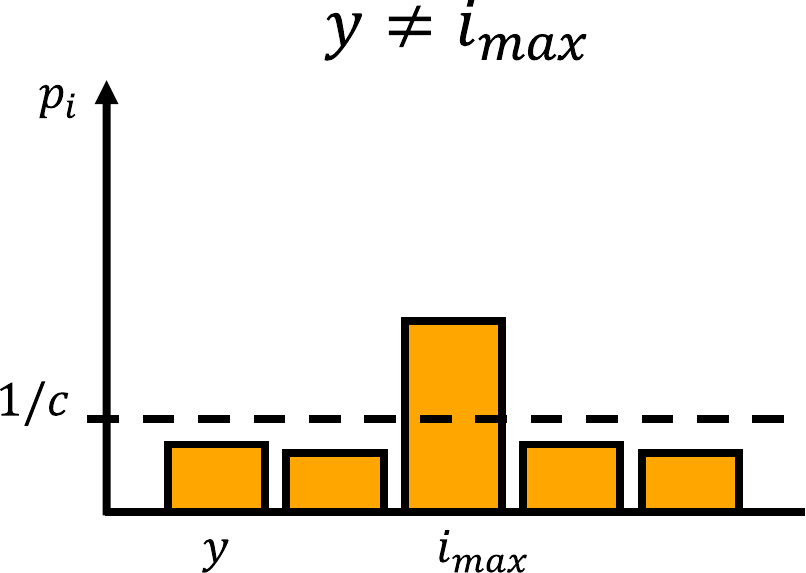} \\
(a) & (b) \\
\\
\includegraphics[width=4cm]{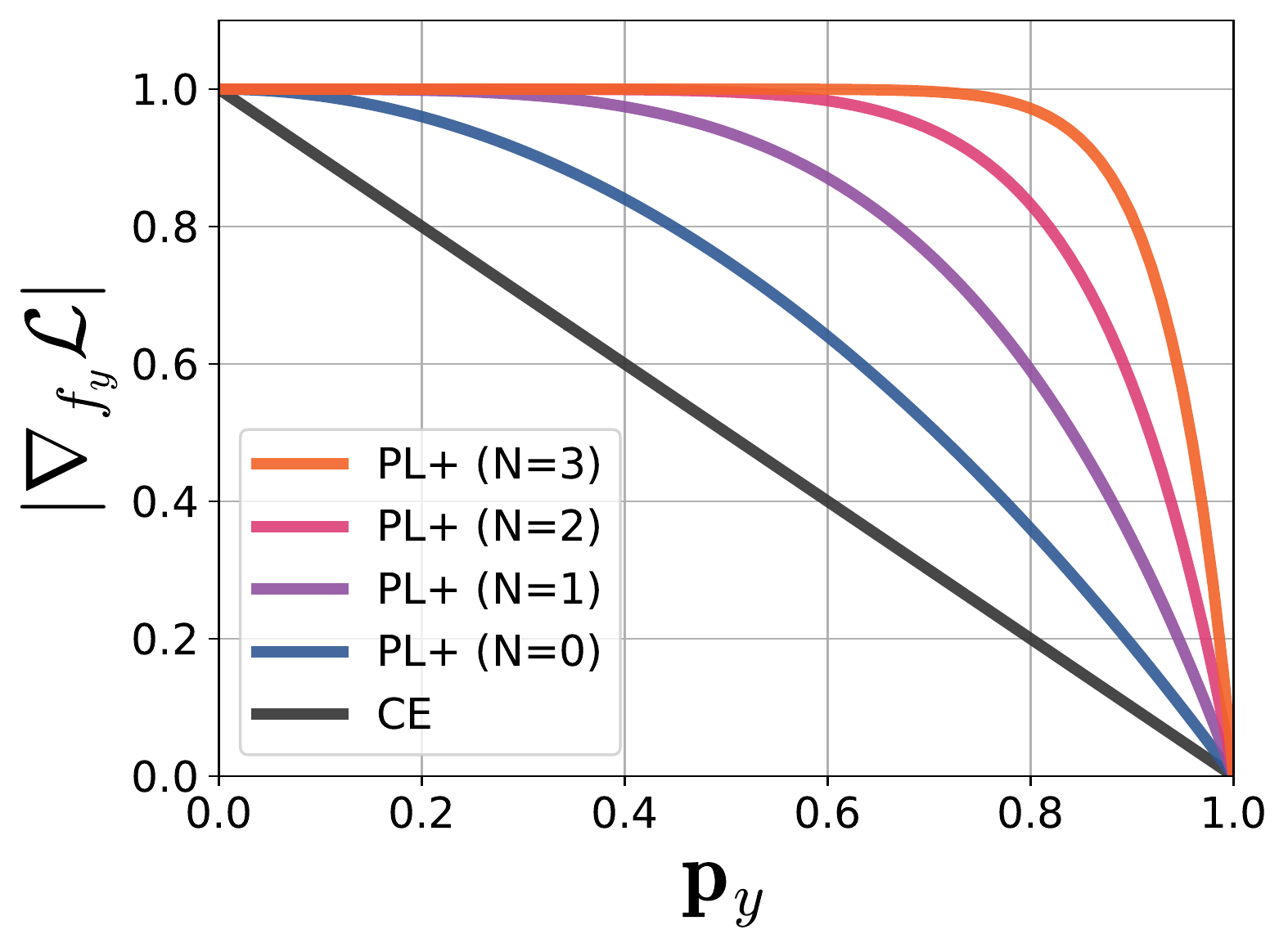} & \includegraphics[width=4cm]{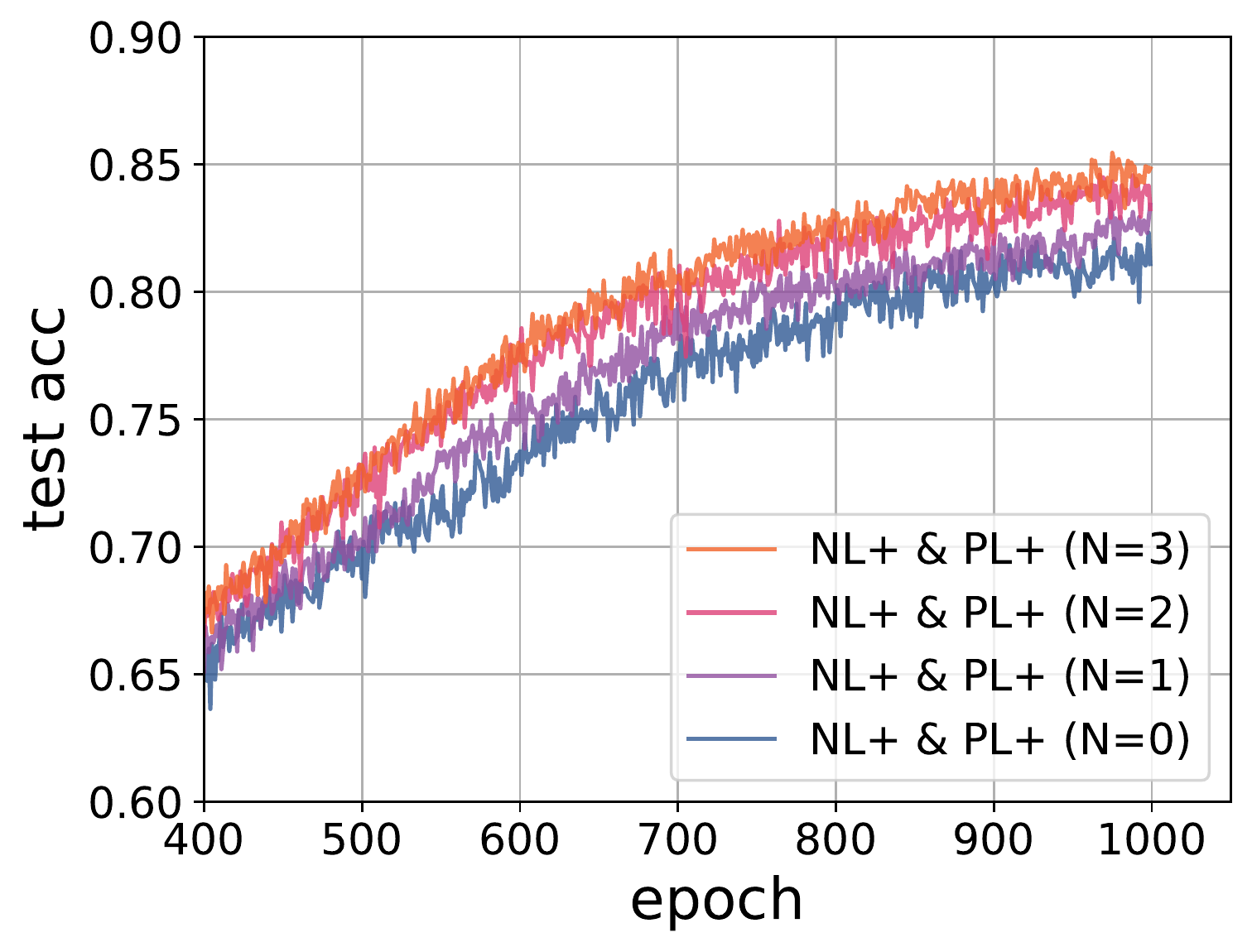} \\
(c) & (d)
\end{tabular}
\caption{(a), (b): Cases for selecting data for PL+. Data is the candidate for PL+ if confidences at classes other than label of maximum probability is under uniform distribution ($1/c$). (c): Gradient of PL+ depending on $N$ compared to PL (cross-entropy loss). (d): Accuracy comparison between PL+ with different $N$. This shows that the flatter version of PL+ ($N=3$) generates better training results.}
\label{fig:SelPL_SelPL+}
\vspace{-5mm}
\end{figure}

\subsection{PL+}
\label{sec:PL+}
In this section, we introduce the second loss function $\mathcal{L}_{PL+}$ in JNPL. As mentioned in Section~\ref{sec:Introduction}, when training data is verified to have clean labels, PL is a faster and more accurate method than NL. Following this fact, we apply PL+ to our method for faster convergence. But compared to NLNL, this is not applied in a sequential step but rather as a unified step.

First of all, the criteria for selecting the training data for PL+ is required. Previously, NLNL applied PL to data over the threshold ($\boldsymbol{p}_{y}>0.5$). However, the criteria for selecting data for PL should be stricter. Even if a data satisfies $\boldsymbol{p}_{y}>0.5$, a probability of other class may reach as much as 0.5, resulting in the risk of selecting noisy data as clean data. Hence, PL+ considers the probabilities of classes other than the given label. When probabilities of other classes except for given label are under uniform distribution $\frac{1}{c}$, this data is a candidate for PL+ (Figure~\ref{fig:SelPL_SelPL+} (a)). Additionally, among the candidates for PL+, it is selected through Bernoulli sampling with respect to $\boldsymbol{p}_{y}$. The higher the $\boldsymbol{p}_{y}$, the more frequently the data would be trained with PL+. Furthermore, PL+ selects data not only from expected-to-be-clean data but also from noisy data. Meaning that, when the probabilities of other classes except for the label of maximum probability is under the uniform distribution, the data is also a candidate for PL+ using the maximum probability class label ($=\hat{y}$) (Figure~\ref{fig:SelPL_SelPL+} (b)). In this way, PL+ selects data for training more strictly, but also, the candidate area is increased. The pseudocode for PL+ process is shown in Algorithm~\ref{alg:PL+ candidate}.

\begin{algorithm}[t]
\SetAlgoLined
\KwIn{mini-batch $\mathcal{\bar{D}}$}
\KwResult{$\mathcal{L}_{PL+}$ over mini-batch $\mathcal{\bar{D}}_{PL+}$}
\For{$(\boldsymbol{x}, y) \in \mathcal{\bar{D}}$}{
    $\boldsymbol{p} \leftarrow softmax\left(f(\boldsymbol{x})\right)$\\
    $\hat{y} \leftarrow \operatorname*{argmax}_i \boldsymbol{p}_{i}$ \\
    \eIf{$\boldsymbol{p}_i < \frac{1}{c}$ for $\forall i \in \{1,...,c\}\setminus\{\hat{y}\} $}{
        Append $ \left( \boldsymbol{x},\hat{y} \right) $ to $\mathcal{\bar{D}}_{PL+}$ with probability $\boldsymbol{p}_{\hat{y}}$\\
    }{
        Reject  $\left( \boldsymbol{x},y \right) $\\
    }
}
Calculate $ \mathcal{L}_{PL+}\left( f(\boldsymbol{x}),\hat{y} \right)$ for $\mathcal{\bar{D}}_{PL+}$ by Eq.~(\ref{eq:PL+})\\

\KwRet{$\frac{1}{|\mathcal{\bar{D}}_{PL+}|}\sum_{x\in \mathcal{\bar{D}}_{PL+}} 
\mathcal{L}_{PL+}\left( f(\boldsymbol{x}),\hat{y} \right) $}\\
 \caption{PL+}\label{alg:PL+ candidate}
\end{algorithm}

PL is usually done using cross-entropy (CE) loss (Eq~\ref{eq:PL}). However, while it may be tolerable when training clean data, it may not be as tolerable as when training noisy data. The reason for PL in our method is to train faster on more confident data. However, when observing the gradient of CE in Figure~\ref{fig:SelPL_SelPL+} (c), it states that a smaller gradient is provided to more confident data, while a higher gradient is provided to less confident data. Since the goal is to train faster on more confident data, not just training more on less confident data, we propose PL+ loss function to resolve this issue as follows: 
\begin{equation}
\label{eq:PL+}
\mathcal{L}_{PL+}(f,\hat{y}) = -\prod_{n=0}^{N}(1+\boldsymbol{p}_{\hat{y}}^{2^n})\sum_{k=1}^c \boldsymbol{y}_k \log \boldsymbol{p}_k,
\end{equation}
and the gradient of PL+ loss is as follows:
\begin{equation}
\label{eq:PL+ gradient}
\begin{split}
\nabla{\mathcal{L}_{PL+}}&=\prod_{n=0}^{N}(1+\boldsymbol{p}_{\hat{y}}^{2^n})\nabla{\mathcal{L}_{PL}}\\&=-\prod_{n=0}^{N}(1+\boldsymbol{p}_{\hat{y}}^{2^n})(1-\boldsymbol{p}_{\hat{y}})=-(1-\boldsymbol{p}_{\hat{y}}^{2^{N+1}}).
\end{split}
\end{equation}
Similar to NL+, $\prod_{n=0}^{N}(1+\boldsymbol{p}_{\hat{y}}^{2^n})$ acts as a constant weighting factor. By applying this weight factor, the gradient of PL+ loss function is modified as shown in Eq~\ref{eq:PL+ gradient} and visualized in Figure~\ref{fig:SelPL_SelPL+} (c). It can be seen that higher gradient is being provided to data of high $\boldsymbol{p}_y$ as $N$ increases. Figure~\ref{fig:SelPL_SelPL+} (d) proves faster convergence as $N$ increases. We set $N=3$ throughout the whole paper. 

\begin{figure}[t]
\begin{center}
\includegraphics[width=6cm]{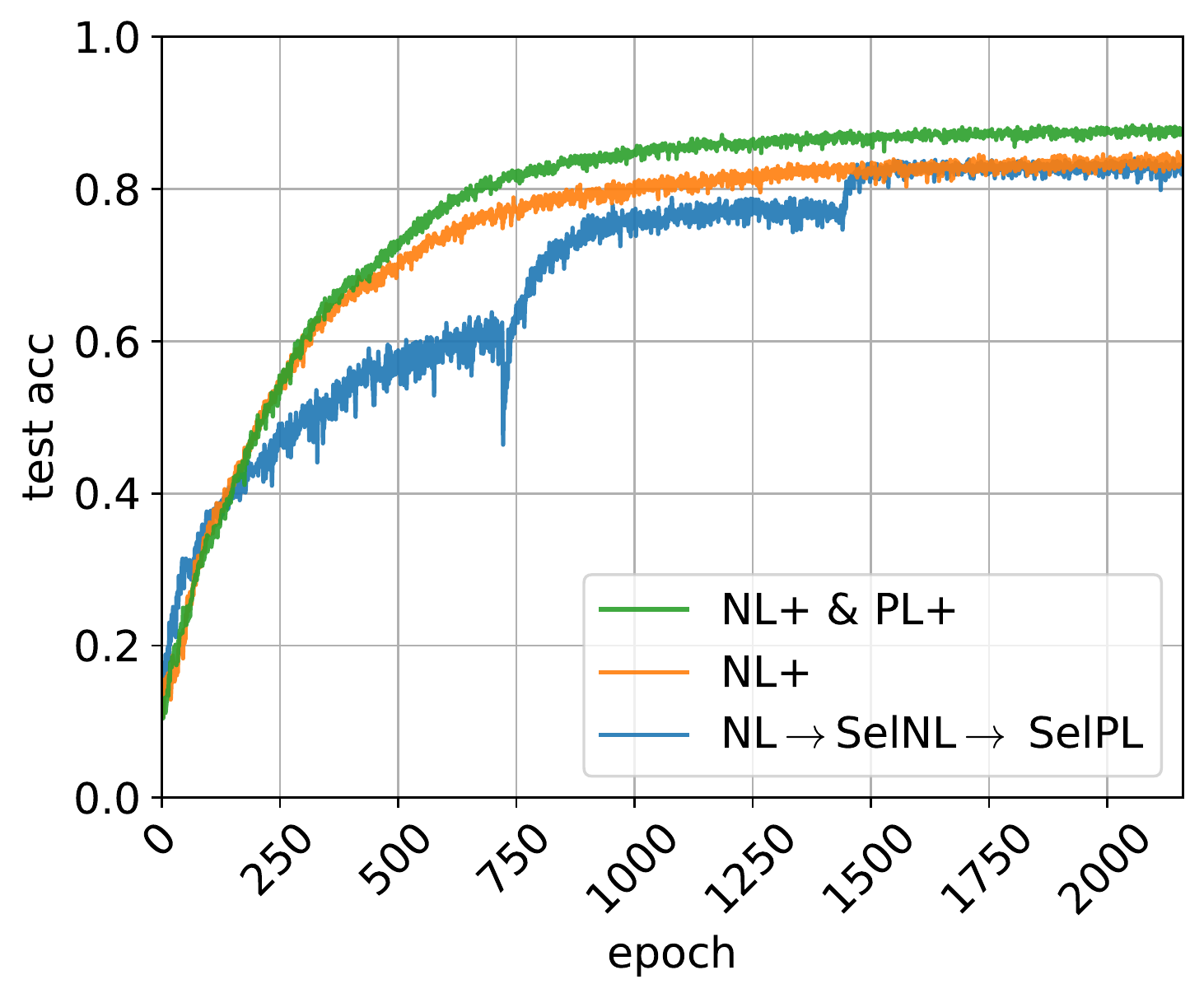} \\
\end{center}
\caption{Accuracy graph of NLNL, NL+, and JNPL (NL+ \& PL+) with CIFAR10 mixed with 60\% \textit{symm} noise.}
\label{fig:SelNLPL_vs_NL+SelPL+_acc}
\vspace{-5mm}
\end{figure}

\section{Analysis}
\label{sec:Analysis}

Since our method is the advanced version of NLNL, which is targeted throughout our whole paper, this section further demonstrates the distinct advantage of our method JNPL over NLNL. 

First of all, our method JNPL is a unified step pipeline for filtering noisy data, compared to 3-step pipeline of NLNL. JNPL is trained with two loss functions simultaneously, increasing the efficiency of training CNN. Figure~\ref{fig:SelNLPL_vs_NL+SelPL+_acc} shows the performance comparison between NLNL (NL$\rightarrow$SelNL$\rightarrow$SelPL), NL+, and JNPL (NL+\&PL+) when training with CIFAR10 mixed with 60\% \textit{symm} noise. Figure~\ref{fig:SelNLPL_vs_NL+SelPL+_acc} clearly indicates that NL+ solely reaches the accuracy of NL$\rightarrow$SelNL, proving better convergence of NL+ compared to NL. Furthermore, when PL+ is done simultaneously along with NL+, it results in faster training without the need for additional subsequent step. It also shows overall accuracy of NL+ and JNPL overpasses the accuracy reached by NLNL while preventing overfitting to noisy data, proving the superiority of our method over NLNL.

\begin{figure*}
\begin{center}
\begin{tabular}{ccc}
\includegraphics[width=4cm]{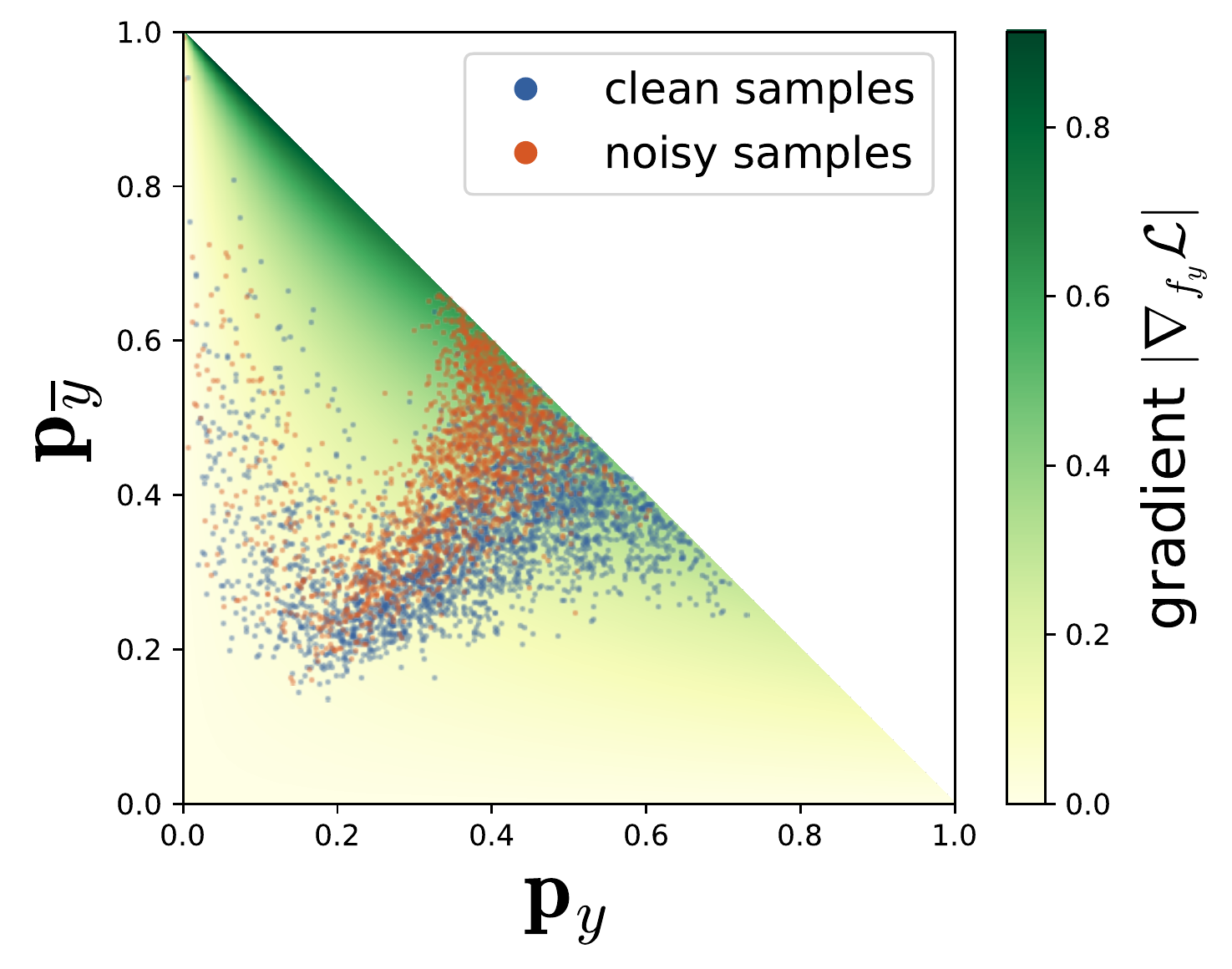}&\includegraphics[width=4cm]{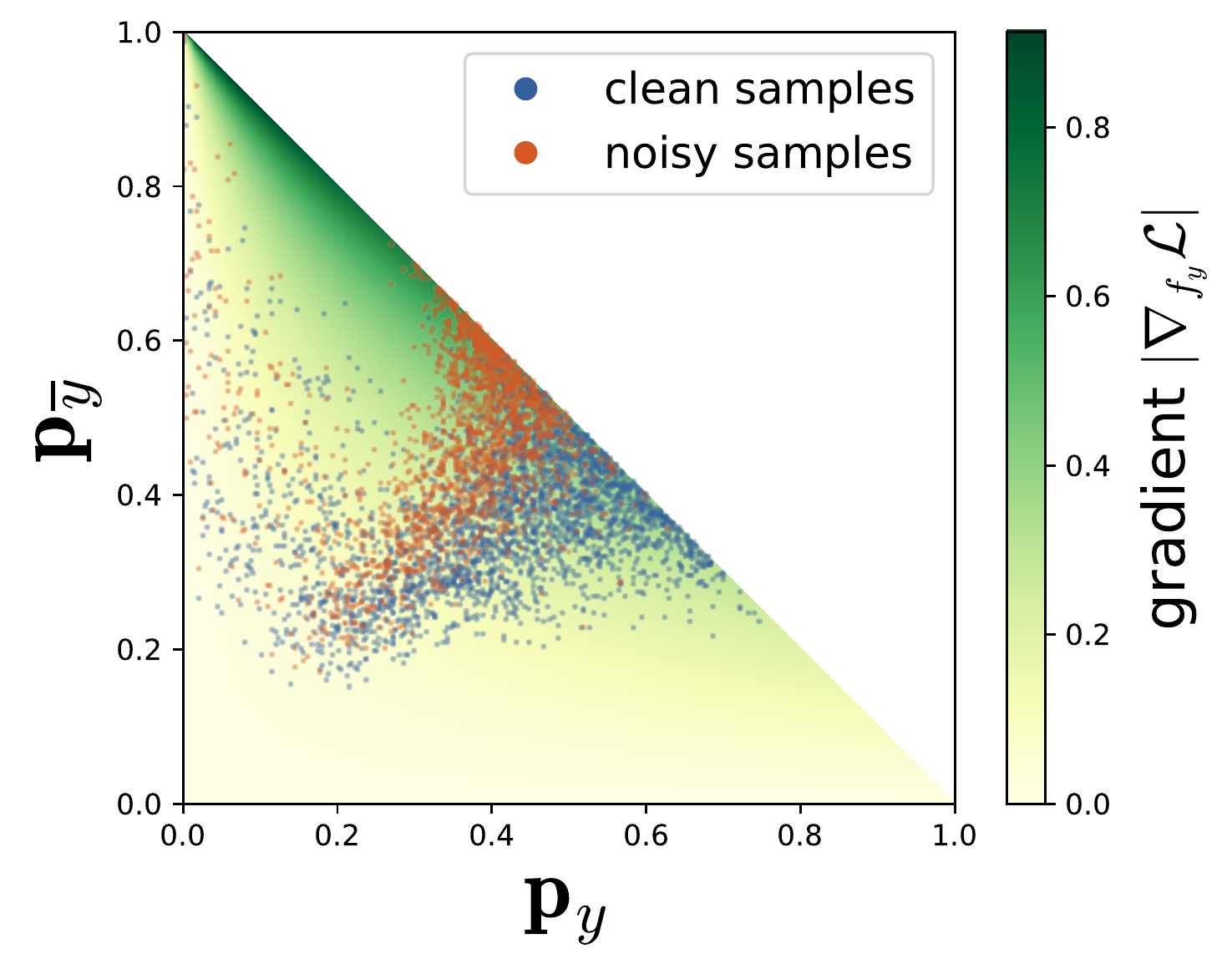}&\includegraphics[width=4cm]{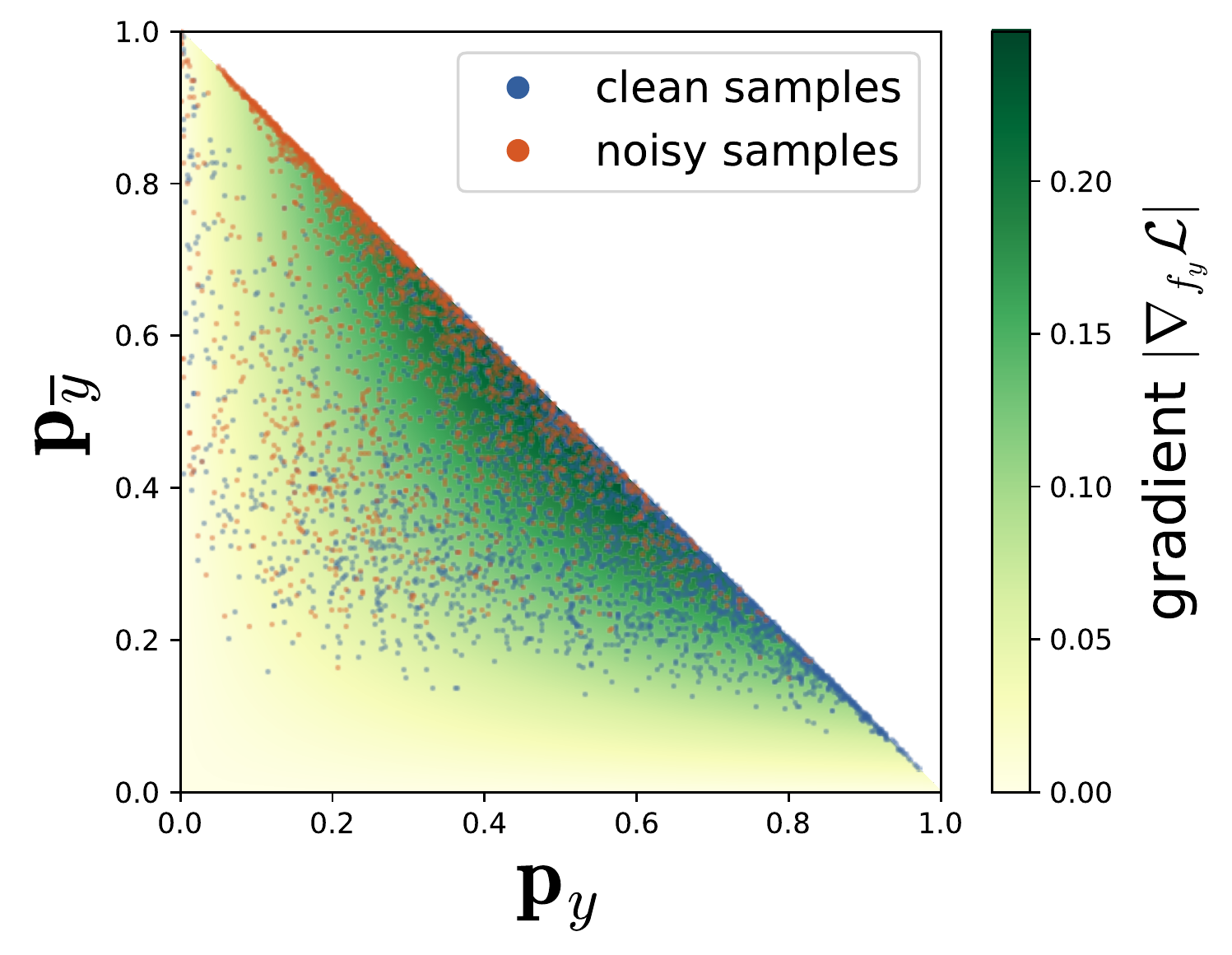}\vspace{-1.1mm}\\
(a) & (b) & (c) \vspace{1.1mm}\\
\includegraphics[width=4cm]{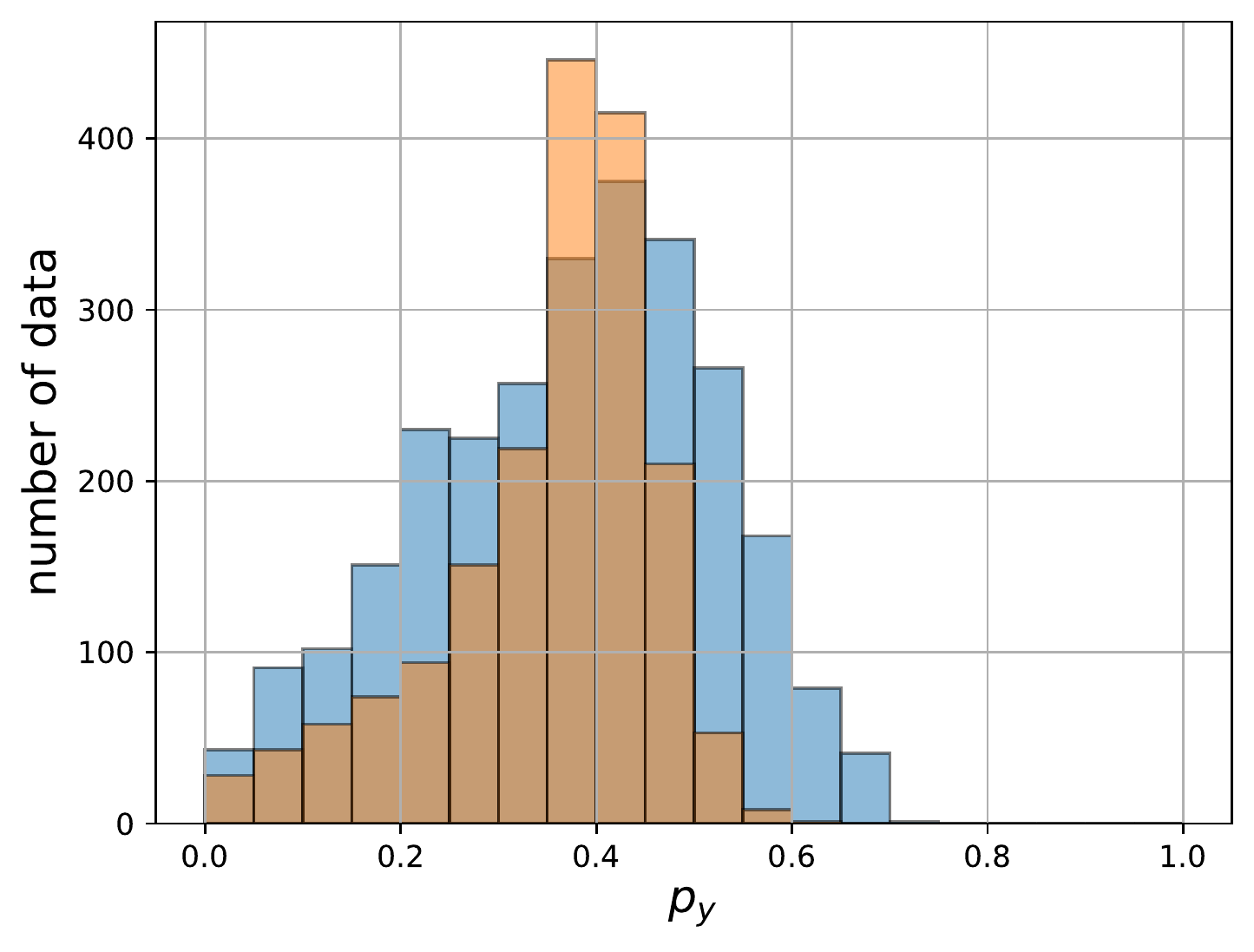}&\includegraphics[width=4cm]{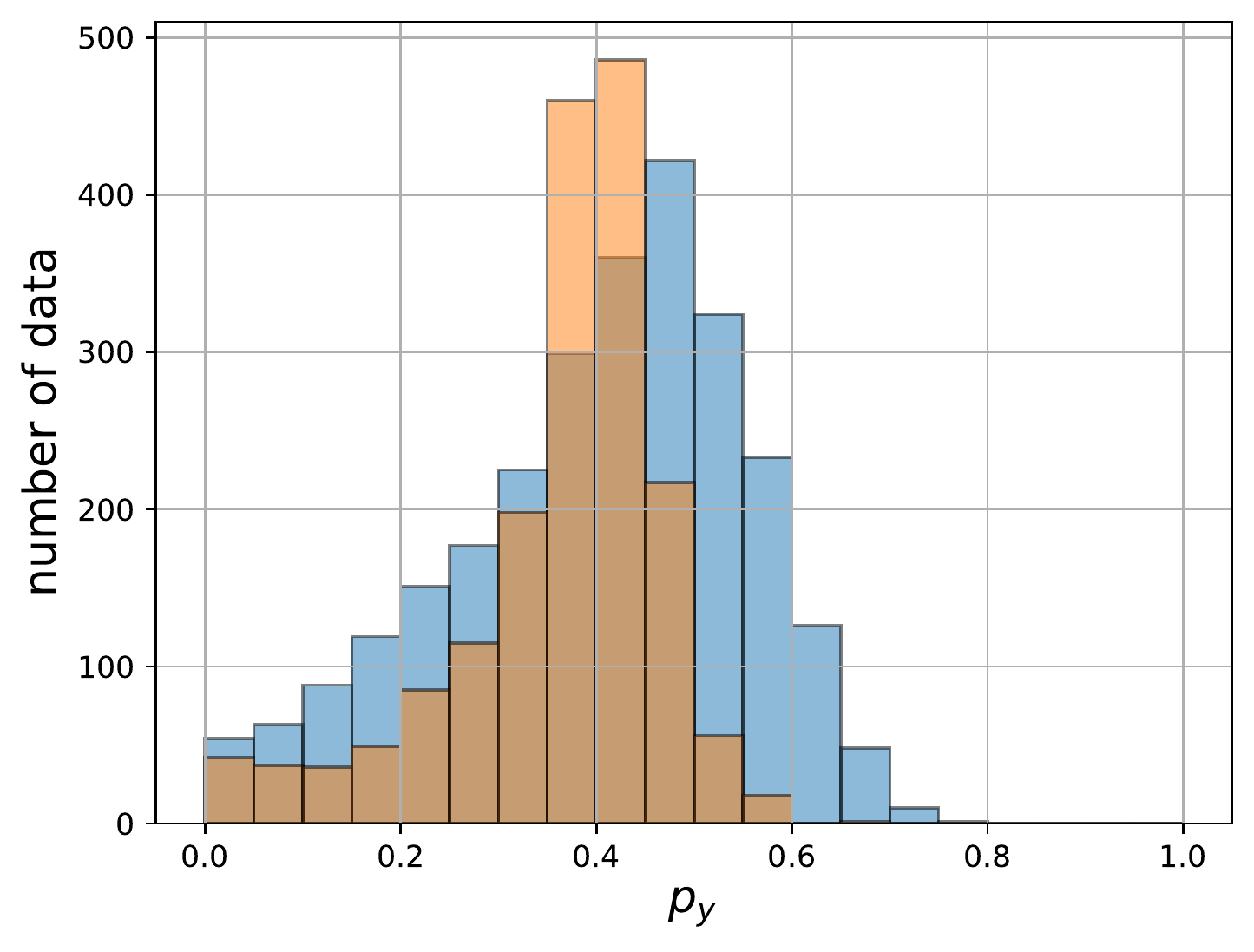}&\includegraphics[width=4cm]{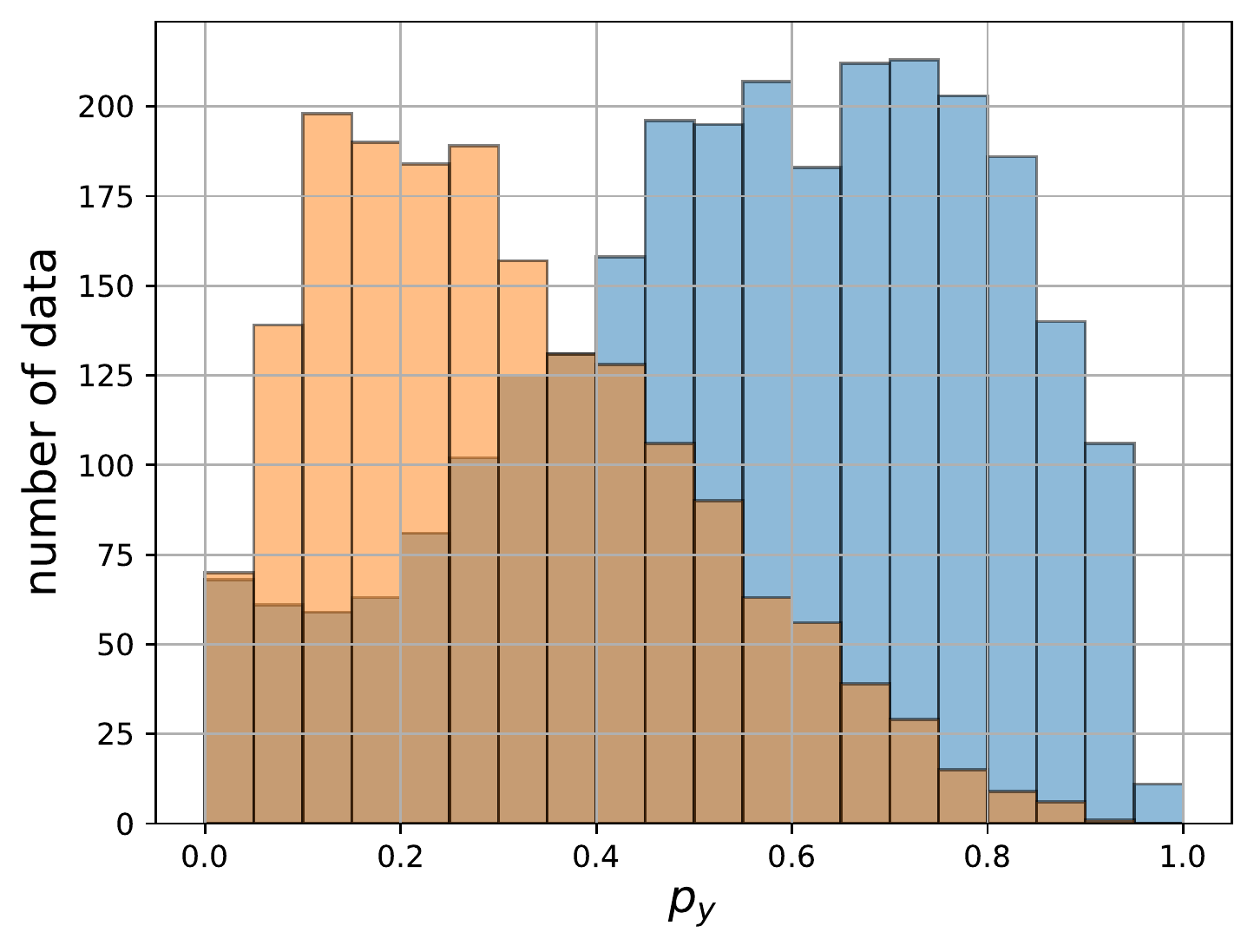}\vspace{-1.1mm}\\
(d) & (e) & (f)
\end{tabular}
\caption{Comparison between NL and NL+ for \textit{asymm} 40\% noise CIFAR10 ``cat'' class. (a), (d): Gradient map and histogram of NL, respectively. (b), (e): Gradient map and histogram of NL$\rightarrow$SelNL, respectively. (c), (f): Gradient map and histogram of NL+, respectively. Blue indicates clean data whereas orange indicates noisy data in histograms.}
\label{fig:NL_NL+_asymm40}
\end{center}
\vspace{-5mm}
\end{figure*}

Secondly, NL+ is more capable of handling more diverse noise types compared to NL$\rightarrow$SelNL owing to the nature of gradient followed by $\mathcal{L}_{NL+}$. Although NL applies SelNL to compensate for underfitting problem, we show that this is not an optimal solution for all types of noise. Consider when training data is CIFAR10 mixed with \textit{asymm} noise, especially when class ``dog'' is mixed with ``cat'' in bidirectional manner (DOG $\leftrightarrow$ CAT). Overall probability values across all classes are shared between class ``dog'' and ``cat,'' resulting in distribution of training data as shown in Figure~\ref{fig:NL_NL+_asymm40} (a), (d). In this case, SelNL shows almost no effect as the noisy data is not under the uniform distribution (Figure~\ref{fig:NL_NL+_asymm40} (b), (e)). Whereas for NL+, due to the fact that gradient for region ($\boldsymbol{p}_y<0.5$ \& $\boldsymbol{p}_{\overline{y}}>0.5$) is reduced in a smooth manner compared to NL, it eventually enables both classes to be separated, showing distinct advantage of NL+ over SelNL (Figure~\ref{fig:NL_NL+_asymm40} (c), (f)).

\begin{figure}[t]
\begin{tabular}{cc}
\includegraphics[width=4cm]{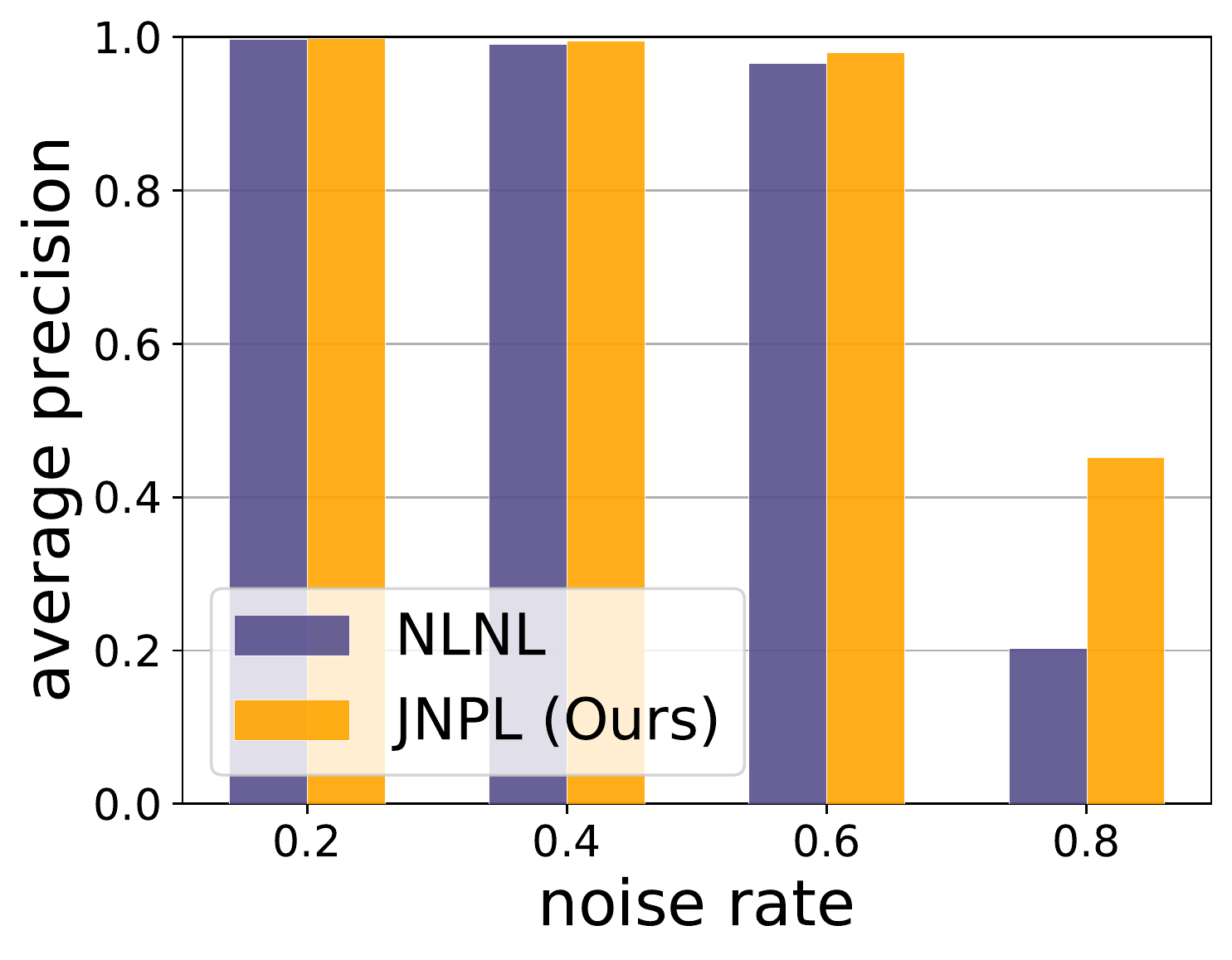} & \includegraphics[width=4cm]{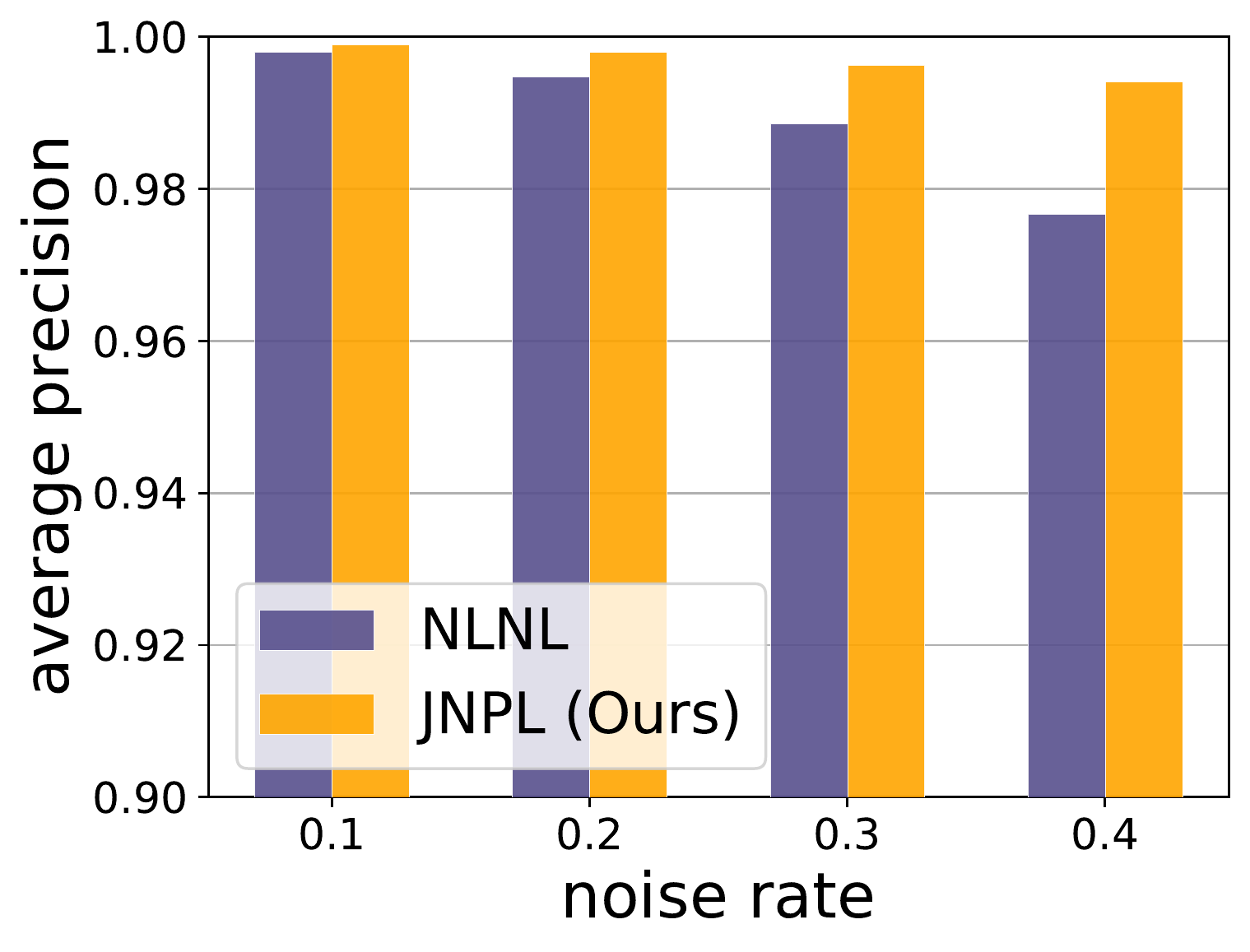} \\
(a) & (b) \\
\\
\includegraphics[width=4cm]{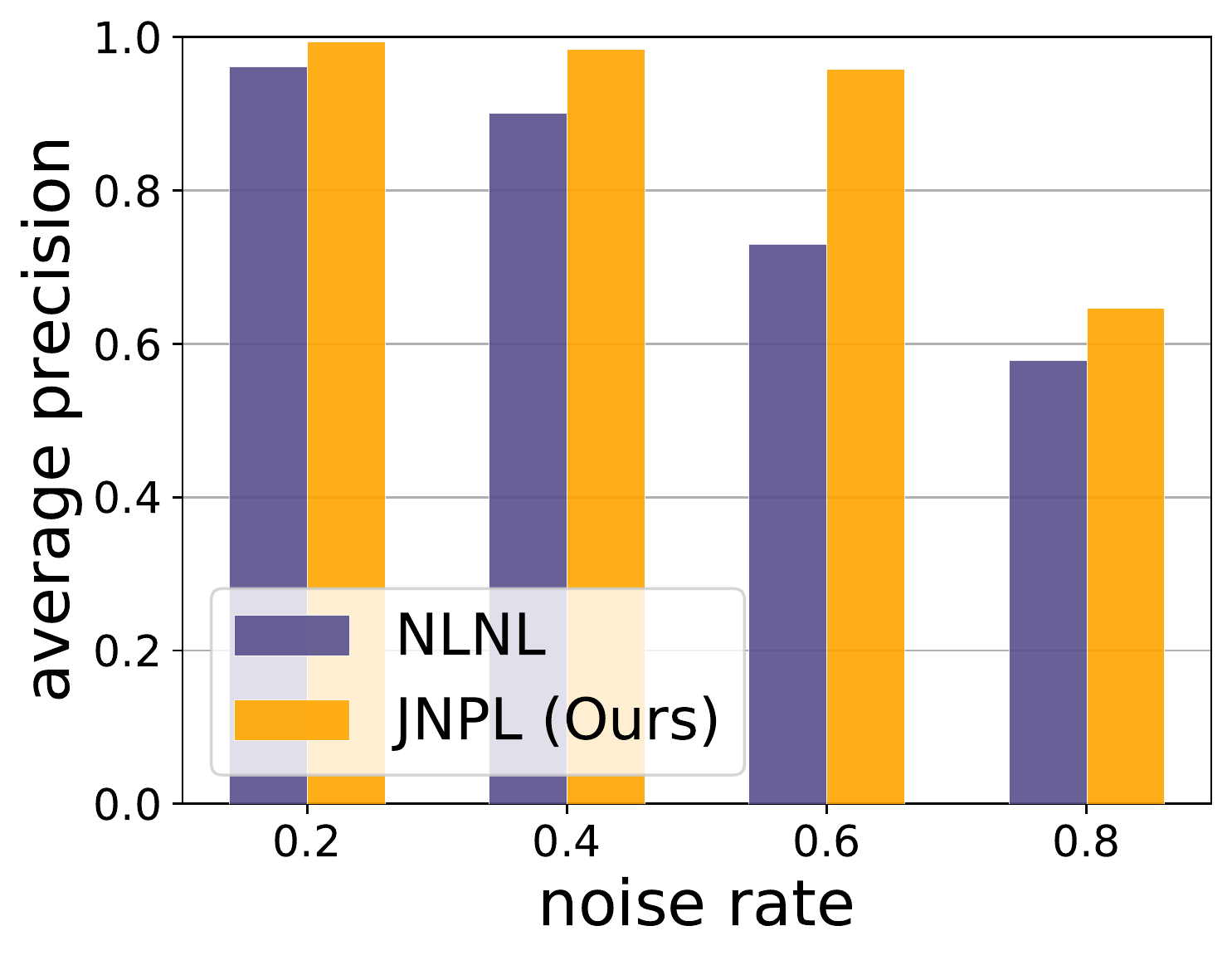} & \includegraphics[width=4cm]{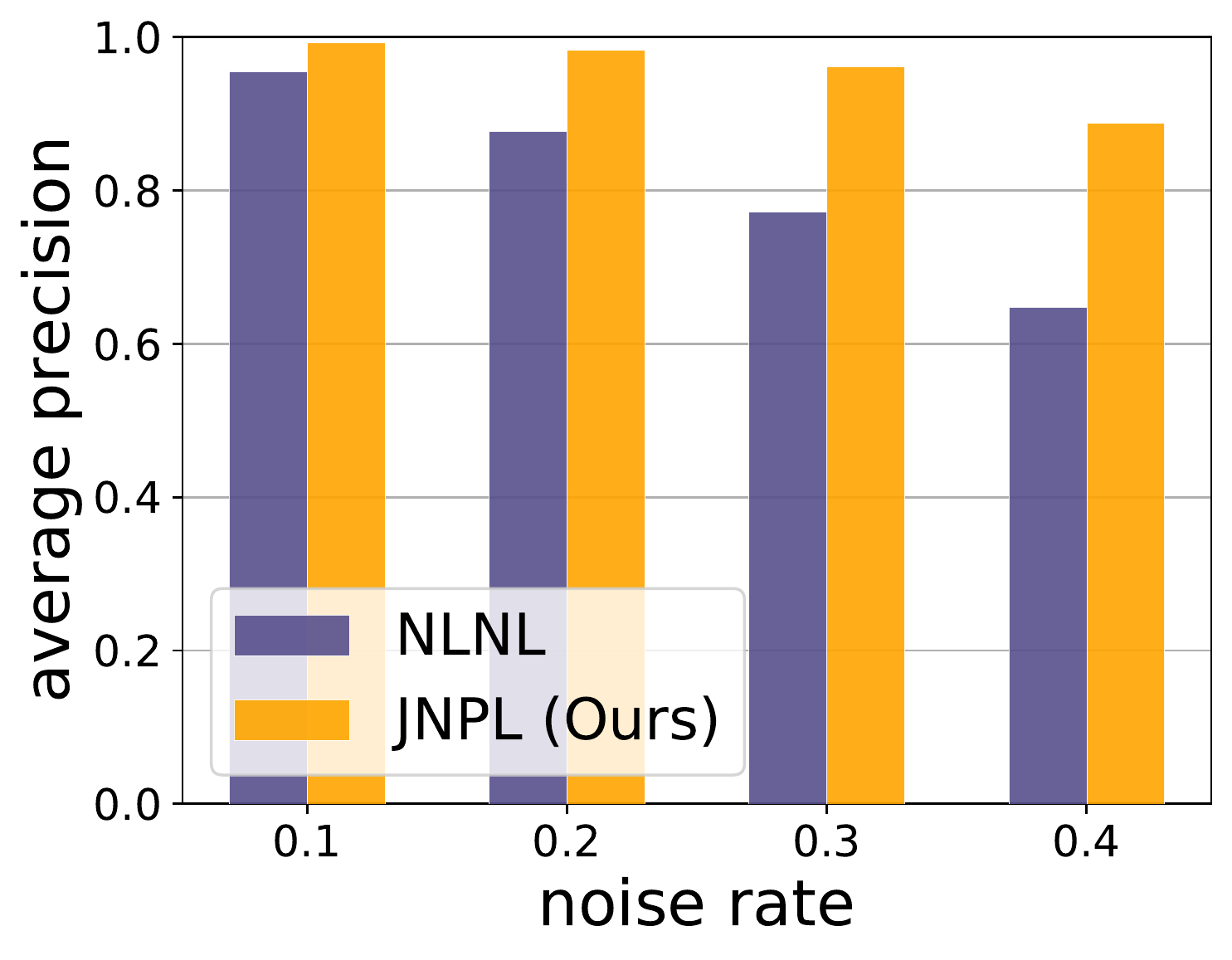} \\
(c) & (d)
\end{tabular}
\caption{Average Precision (AP) for CIFAR10 / CIFAR100 on \textit{symm} / \textit{asymm} noises. (a), (b): AP for CIFAR10 on \textit{symm} / \textit{asymm} noises, respectively. (c), (d): AP for CIFAR100 on \textit{symm} / \textit{asymm} noises, respectively.}
\label{fig:average_precision}
\vspace{-5mm}
\end{figure}

Finally, we show that our method JNPL successfully filters noisy data from training data than NLNL. Figure~\ref{fig:average_precision} shows overall filtering ability between NLNL and JNPL with average precision (AP). It is compared in diverse environment: CIFAR10/CIFAR100 mixed with different ratio of \textit{symm} and \textit{asymm} noise. It shows that our method outperforms NLNL in filtering noisy data on overall cases. Furthermore, it can be observed that 
gap of AP between NLNL and JNPL increases as the noise ratio increases. This implies that JNPL is more robust to the amount of noise mixed in training data. Also, JNPL being more robust to \textit{asymm} noise than NLNL also proves the point made above. This phenomenon is more clearly shown in more difficult data CIFAR100. AP of NLNL drastically decreases as the noise rate gets higher. However, JNPL shows robustness in types and ratios of noise, similar to when training with CIFAR10. Figure~\ref{fig:average_precision} demonstrates our method JNPL is capable of being generalized to type and ratio of noise, and even number of classes in the dataset.

\section{Experiments}
\label{sec:Experiments}

In this section, we describe the experiments performed to evaluate our method. Pseudo-labeling is done on a training dataset filtered by JNPL for noisy data classification and resulting accuracies are compared to those of other existing methods. We verify our method by comparing with other recent baseline methods, varying experimental settings in terms of dataset and type and ratio of noise in the training data.

\subsection{Experiment settings}
\label{sec:Experiment settings}
\noindent\textbf{Baseline methods} We compare our method against CE, along with recent state-of-the-art approaches including Co-teaching~\cite{han2018co}, JoCoR~\cite{wei2020combating}, APL~\cite{ma2020normalized}, and NLNL~\cite{kim2019nlnl}.

\noindent\textbf{Dataset} We conduct the experiments on CIFAR10, CIFAR100 \cite{cifar-10} mixed with two types noises (\textit{symm}, \textit{asymm}), and Clothing1M  \cite{xiao2015learning} dataset. Clothing1M is a large-scale real-world dataset with noisy labels, containing 1 million images of clothing obtained from several online shopping websites. It is reported that the overall accuracy of noisy labels in this dataset is 61.54\%, and some pairs of classes are often confused with each other (\textit{e.g.}, Knitwear and Sweater). For preprocessing, we performed mean subtraction, horizontal flip, and random crops for CIFAR10 and CIFAR100. For Clothing1M, we resize the image to 256$\times$256, crop 224$\times$224 at the center and perform mean subtraction and horizontal flip.

\noindent\textbf{Label noise types} We generated noisy CIFAR10 and CIFAR100 datasets according to the following procedures. In symmetric (\textit{symm}) noise experiments, we flipped a portion of the labels by re-sampling each label uniformly from the remaining classes, excluding the ground-truth class. In asymmetric (\textit{asymm}) noise experiments, we followed the same label transition rule used by Patrini \etal~\cite{patrini2017making}. For CIFAR10, we mapped TRUCK $\rightarrow$ AUTOMOBILE, BIRD $\rightarrow$ PLANE, DEER $\rightarrow$ HORSE, and CAT $\leftrightarrow$ DOG. For CIFAR100, the noise flipped each class into the next, circularly within super-classes.

For each noise type, we compared the methods under the symmetric noise rates of $\eta_{symm} \in \{0.2, 0.4, 0.6, 0.8 \}$ and asymmetric noise rates of $\eta_{asymm} \in \{0.1, 0.2, 0.3, 0.4 \}$.

\noindent\textbf{Models} For CIFAR10 and CIFAR100 experiments, we used ResNet34. For Clothing1M, we used ResNet50~\cite{he2016deep}, pre-trained on ImageNet.

\noindent\textbf{Hyperparameters} We used stochastic gradient descent (SGD) with momentum of 0.9, weight decay of $10^{-4}$. For experiments with CIFAR10 and CIFAR100, batch size is set to 128. Moreover, JNPL trains CNN for 1000 epochs with initial learning rate of $10^{-2}$, and decay by a factor of 10 at 800 epochs. For pseudo labeling, initial learning rate is 0.1, decayed by a factor of 10 at 192, 288 epochs (480 epochs total). For experiments with Clothing1M, batch size is set to 64, and JNPL trains CNN for 40 epochs with initial learning rate of $10^{-3}$, and decay by a factor of 10 at 30 epochs. For pseudo labeling, initial learning rate is $10^{-3}$, decayed by a factor of 10 at 10 epochs (15 epochs total).

For CIFAR100, we adopt the technique NLNL proposed for generalization to the number of classes in training data: providing multi $\overline{y}$ to each data. We provide 110 $\overline{y}$ to each data in order to match the training speed to when training with CIFAR10~\cite{kim2019nlnl}.

\begin{table*}
\begin{center}
\begin{tabular}{c|c|c|cccc|cccc}\toprule
\multirow{2}{*}{Datasets} & \multirow{2}{*}{Model} & \multirow{2}{*}{Methods} & \multicolumn{4}{c|}{\textit{Symm}} & \multicolumn{4}{c}{\textit{Asymm}} \\
 & & & 20 & 40 & 60 & 80 & 10 & 20 & 30 & 40 \\
\midrule
\multirow{7}{*}{CIFAR10} & \multirow{7}{*}{ResNet34} & CE & 83.95 & 67.58 & 43.55 & 17.32 & 91.39 & 87.67 & 82.73 & 76.37\\
 & & Co-teaching~\cite{han2018co} & 91.08 & 88.08 & 80.96 & 21.13 & 94.20 & 93.24 & 90.67 & 70.20\\
 & & JoCoR~\cite{wei2020combating} & 91.84 & 88.15 & 59.20 & 20.72 & 93.13 & 91.19 & 89.01 & 83.61\\
 & & NFL+RCE~\cite{ma2020normalized} & 90.50 & 85.16 & 70.77 & 19.67 & 92.35 & 89.66 & 84.92 & 78.30\\
 & & NCE+RCE~\cite{ma2020normalized} & 90.36 & 84.57 & 74.09 & \textbf{26.71} & 91.89 & 90.13 & 85.80 & 78.49\\
 & & NLNL~\cite{kim2019nlnl} & \textbf{94.23} & \textbf{92.43} & \textbf{88.32} & - & \textbf{94.57} & \textbf{93.35} & \textbf{91.80} & \textbf{89.86}\\
 & & Ours & \textbf{93.53} & \textbf{91.89} & \textbf{88.45} & \textbf{35.65} & \textbf{94.22} & \textbf{93.45} & \textbf{92.47} & \textbf{90.72}\\
\midrule
\multirow{7}{*}{CIFAR100} & \multirow{7}{*}{ResNet34} & CE & 57.32 & 45.64 & 24.30 & 8.06 & 65.12 & 62.12 & 52.77 & 44.55\\
 & & Co-teaching~\cite{han2018co} & 69.56 & 62.81 & 51.12 & 10.25 &\textbf{ 72.52} & \textbf{67.46} & \textbf{61.50} & 52.86\\
 & & JoCoR~\cite{wei2020combating} & \textbf{71.75} & 63.96 & 37.84 & 7.32& 72.01 & 65.05 & 56.63 & 45.14\\
 & & NFL+RCE~\cite{ma2020normalized} & 58.70 & 42.76 & 24.77 & \textbf{10.57} & 63.70 & 56.45 & 46.96 & 37.52\\
 & & NCE+RCE~\cite{ma2020normalized} & 57.41 & 43.75 & 25.87 & 9.94 & 64.24 & 56.48 & 47.17 & 36.40\\
 & & NLNL~\cite{kim2019nlnl} & \textbf{71.52} & \textbf{66.39} & \textbf{56.51} & - & 70.35 & 63.12 & 54.87 & \textbf{45.70}\\
 & & Ours & 70.94 & \textbf{68.11} & \textbf{61.26} & \textbf{17.55} & \textbf{72.03} & \textbf{69.95} & \textbf{68.12} & \textbf{59.51}\\
\bottomrule
\end{tabular}
\end{center}
\caption{Comparison with other baseline methods on CIFAR10, CIFAR100 mixed with various types and ratios of noise. Best 2 accuracies are \textbf{bold faced}.}
\label{tab:Experiment results}
\end{table*}

\begin{table}[]
\centering
\begin{tabular}{l|c}
\toprule
Method & Test Accuracy \\ \midrule
CE & 69.21 \\
Forward~\cite{patrini2016making} & 69.84 \\
M-correction~\cite{arazo2019unsupervised} & 71.00 \\
LIMIT~\cite{harutyunyan2020improving} & 71.39 \\
Joint-Optim~\cite{tanaka2018joint} & 72.16 \\
MetaCleaner~\cite{zhang2019metacleaner} & 72.5 \\
MLNT~\cite{li2019learning} & 73.47 \\
PENCIL~\cite{yi2019probabilistic} & 73.49 \\ \midrule
Ours & \textbf{74.15} \\
\bottomrule
\end{tabular}
\caption{Comparison on Clothing1M with other baseline methods.}
\label{tab:Clothing1M}
\vspace{-3mm}
\end{table}

\subsection{Results}
\label{sec:Results}

Table~\ref{tab:Experiment results} shows the results of our method and other baseline methods in various noise environment and two datasets. Our proposed method outperformed all other comparable baseline methods in overall noise types and ratios. The result shows other baseline methods achieve comparable results in the less-noisy environment, but the performance decreases drastically as the noise ratio increases, which is even more visible at CIFAR100, which is the harder case for noisy data classification. Our method shows a distinct improvement in this situation compared to all other methods. It was shown in Section~\ref{sec:Analysis} our method is robust to the amount of noise mixed in training data, regardless of the type of noises. Table~\ref{tab:Experiment results} shows a similar result that our method achieves more distinct best accuracy as the noise rate gets higher. This phenomenon is more emphasized for CIFAR100. Our method outperforms as much as 6 to 7\% at both \textit{symm} and \textit{asymm} noises in this dataset. It is noteworthy that our method achieved 7\% higher state-of-the-art accuracy in the most difficult setting in Table~\ref{tab:Experiment results}, which is 100 class dataset mixed with 40\% \textit{asymm} noise. It is widely known training in general is challenging as the number of classes in the dataset increases. Furthermore, compared to \textit{symm} noise, \textit{asymm} noise is the replica of noise that we can actually make in real-life. Achieving such a high accuracy in this setting implies that our method is more capable of generalizing to training data and various types and ratios of noise mixed within compared to other baseline methods. 

It is shown that Co-teaching and JoCoR method~\cite{han2018co,wei2020combating} exceeds the performance compared to our method for some cases. However, it should be noted that they assume prior knowledge on important statistics about the dataset such as the amount of noise. In reality, this assumption often leaves the method impractical because the ratio of noise mixed in training data is likely to be unknown. On the other hand, our method does not assume any such prior knowledge and therefore does not require extensive tuning of hyper-parameters.

To demonstrate the generalization of our method JNPL to real-world noisy data, we compose an experiment on Clothing1M dataset (Table~\ref{tab:Clothing1M}). We brought recent baseline methods which conducted experiment on Clothing1M for comparison. It shows our method achieves comparable performance, outperforming other recent baseline methods. This result clearly proves that JNPL can generalize to training data mixed with various types and ratios of noise, showing the novelty of our method.


\section{Conclusion}
\label{sec:Conclusion}

We propose Joint Negative and Positive Learning, the next version of NLNL which is the novel single-step pipeline for filtering noisy training data. 
Compared to 3-step pipeline of NLNL, our method trains CNN with two-loss functions ($\mathcal{L}_{NL+}+\mathcal{L}_{PL+}$) in one step.
They are developed from previous NL and PL loss functions to enhance convergence and training speed, resulting in better filtering performance than NLNL. 
We demonstrated that JNPL is stable and robust in various types and ratios of noise mixed in training data. 
Our method achieves state-of-the-art performance in noisy data classification utilizing pseudo-labeling to our filtered training data, proving our method's excellent filtering ability without referring to any prior knowledge. 

{\small
\bibliographystyle{ieee_fullname}
\bibliography{cvpr}
}

\end{document}